\documentclass[review,times,1p]{elsarticle}
\usepackage{siunitx}
\usepackage{lineno,hyperref}
\modulolinenumbers[5]

\journal{ of \LaTeX\ Templates}

\begin{document}

\begin{frontmatter}

\title{Person Identification from Partial Gait Cycle Using Fully Convolutional Neural Network}

\author[mainaddress]{Maryam Babaee\corref{mycorrespondingauthor}}
\author[mainaddress]{Linwei Li}
\author[mainaddress]{Gerhard Rigoll}
\address[mainaddress]{Institute for Human-Machine Communication, Department of Electrical and Computer Engineering, Technical University of Munich, Arcisstr. 21, Munich, Germany}
\cortext[mycorrespondingauthor]{Corresponding author}




\begin{abstract}
Gait as a biometric property for person identification plays a key role in video surveillance and security applications. In gait recognition, normally, gait feature such as Gait Energy Image (GEI) is extracted from one full gait cycle. However in many circumstances, such a full gait cycle might not be available due to occlusion. Thus, the GEI is not complete giving rise to a degrading in gait-based person identification rate. In this paper, we address this issue by proposing a novel method to identify individuals from gait feature when a few (or even single) frame(s) is available. To do so, we propose a deep learning-based approach to transform incomplete GEI to the corresponding complete GEI obtained from a full gait cycle. More precisely, this transformation is done gradually by training several auto encoders independently and then combining these as a uniform model. Experimental results on two public gait datasets, namely OULP and Casia-B demonstrate the validity of the proposed method in dealing with very incomplete gait cycles. 
\end{abstract}

\begin{keyword}
Gait Recognition, Gait Energy Image, Deep Learning, Fully Convolutional Neural Network
\end{keyword}

\end{frontmatter}

\section{Introduction} 
Gait (the style of natural walking)~\cite{Boyd2005} as a biometric property can be used to identify individuals at a distance, when other biometric features such as face, fingerprint, and iris might not be available~\cite{wang2010A}. Here, a sequence of images showing a person walking is analyzed as input data. Since the natural walking of a person is periodic, it is sufficient to consider only one period (gait cycle) from the whole sequence. The gait cycle is defined as the time interval between the exact same repetitive events of walking that generally starts when one foot is in contact with the ground (although any position of the foot during walking can be regarded as the starting point of the gait cycle). 

Since last two decades, many gait-based human identification techniques have been proposed~\cite{xing2016complete,wolf2016multi,Hofmann2013b,babaee2017view}. The main assumption in all these methods is that a full gait cycle of individuals is available, which is a strong assumption in video surveillance applications where occlusion occurs a lot and a person might be observed in only a few frames. From full Gait cycle, a simple and effective gait representation, namely Gait Energy Image (GEI)~\cite{man2006GEI} is computed by averaging of silhouette images of a walking person. This standard gait feature has been widely used alone or in combination with other features in gait recognition systems~\cite{kumar2014lbp, hongye2015gait}. Furthermore, many recent view-invariant gait features have been extracted from the GEI~\cite{zheng2011robust, verlekar2017view, kusakunniran2009multiple, babaee2017view}.

In this paper, we propose a gait-based method to identify a person from a few frames, i.e., incomplete gait cycle. More specifically, having only a few frames of a full gait cycle, we first generate an incomplete GEI. Next, we train a Fully Convolutional Neural Network (FCN) which gets the computed incomplete GEI (average of a few frames) as input and outputs the reconstructed complete GEI. This transformation is done in a progressive way, which means first several auto-encoders are trained as small range regressors. Next the hidden layers of these auto encoders are combined to have a uniform end-to-end network. Here, we have used GEI feature, since it is widely used in many applications. However, the proposed approach can be applied on other gait features such as gait entropy image or gait flow image. The conducted experiments confirm that the proposed network can successfully reconstruct a complete GEI. This can improve gait recognition performance when the input data is not completely available. Therefore, it has great potential in real scenes.

The rest of the paper is organized as follows; Section~\ref{sec:relatedwork} reviews related works. The proposed methodology is explained in Section~\ref{sec:approach}, where the structure of the proposed network is illustrated. Experiments and evaluation results are presented in Section~\ref{sec:experiments}. In the end, Section~\ref{sec:conclusion} provides a short summary and the conclusions.



\section{Related work}
\label{sec:relatedwork} 
Many gait recognition methods have been proposed in the last two decades; some are model-based approaches~\cite{bouchrika2007model,yam2015gait,bobick2001gait, yam2004automated}, while others are appearance-based~\cite{roy2012gait,liu2006improved, wang2010A}. Model-based approaches mainly focus on modeling a person's walking represented by the sticks and the joints. In appearance-based methods, appearance features are extracted from one gait cycle. The gait energy image~\cite{man2006GEI} is the most common gait feature in the latter group. The GEI is generated by simply averaging the gait silhouettes over one gait cycle. Other appearance-based feature representations include chrono-gait image (CGI)~\cite{wang2012human}, gait flow image (GFI)~\cite{lam2011gait}, and gait entropy image (GEnI)~\cite{bashir2009gait}.
The CGI is a temporal template in which the temporal information among gait frames is encoded by a color mapping function. The CGI is obtained by composing the color encoded gait contour images in a gait cycle. The GFI is generated using optical flow to aggregate gray scale contours over one gait cycle. The GEnI represents the randomness of pixel values in the silhouettes image sequence. 

In addition, there are also some gait feature representations based on the GEI including enhanced gait energy image (EGEI)~\cite{chunli2010behavior}, and masked gait energy image (MGEI)~\cite{bashir2010gait}. The EGEI retains the dynamic region information to enhance the GEI feature. The MGEI extracts the dynamic information of a gait sequence based on computed entropy. The MGEI is computed based on the GEI and the entropy of every pixel in order to create a binary feature selection mask of GEI. Iwama et. al~\cite{iwama2012ouisir} show that the GEI is more effective in gait recognition than other gait feature representations. 

The performance of gait recognition is affected by factors such as clothing, carrying objects, view variances, and more importantly, occlusion. Many invariant gait feature representations have been developed, especially for generating view invariant gait features.  

Deep learning-based approaches perform very well in almost every image and video processing application, including gait recognition. Recently, extracting informative and view-invariant gait features using deep learning has gained significant attention. For example, an invariant feature extraction method has been presented in~\cite{yu2017invariant}, where the authors propose to train a convolutional neural network (CNN) using different gait data, including GEIs under different views and different carrying conditions, to extract invariant gait features. These features are learned in a progressive way using multi-stacked auto-encoders. Shiraga et al.~\cite{shiraga2016geinet} propose a CNN called GEINet to extract view-invariant gait features. As a memory-based recognition system, Liu et al.~\cite{liu2016memory} extract 2D positions of joints using the migratory articulated human detection method. Then, a recurrent neural network (RNN) is used for gait recognition. Chao Yan et al.~\cite{yan2015multi} propose to use a CNN and a multi-task learning model (MLT) to identify human gait and to predict multiple human attributes (like view and walking condition) simultaneously. Munif Alotaibi et. al~\cite{alotaibi2017improved} present a gait recognition method using a specialized CNN. They developed a specialized deep CNN architecture that consists of multiple convolutional and sub-sampled layers. This structure is insensitive to earlier mentioned co-variant factors which lead to degradation of recognition rates. 
All mentioned methods compute gait features from one full gait cycle. According to our knowledge, no approach has addressed the problem of gait recognition from an incomplete gait cycle yet.

A fully convolutional network (FCN) is a type of deep neural network which can perform image-to-image transformation. It has been successfully applied in image and video super resolution~\cite{sajjadi2017enhancenet, kappeler2016video}, semantic segmentation~\cite{long2015fully}, and object detection~\cite{dai2016r}. Dong et al.~\cite{dong2016image} propose a FCN for single image super resolution. This approach can learn an end-to-end mapping between a low resolution image and a high resolution image, and outperforms traditional sparse-coding-based single image super-resolution methods. Our proposed model for incomplete GEI to complete GEI transformation is based on FCN architecture.  
   
\section{Approach}
\label{sec:approach} 
Our proposed approach for identification of individuals from incomplete gait cycle is accomplished in two steps. First, an incomplete  GEI is computed from available silhouette images by simply averaging them. Second, we reconstruct a complete GEI (RC-GEI) from the incomplete GEI (IC-GEI) using the proposed model. 

An incomplete GEI for a specific person varies based on the number of frames and the starting point from which it is generated. This means, in an extreme case, there might be only one or two frames of a gait cycle available. In such case, handling the large difference between input (incomplete GEI) and target (complete GEI) is not easy for a single auto-encoder. Thus, we train an end-to-end auto-encoder called ITCNet (incomplete to complete GEI network) to do this transformation gradually. More specifically, this end-to-end network is composed of the hidden layer of $9$ FCNs, each responsible for mapping in a small range; $1/10$ of the gait cycle interval. For instance, if the gait cycle is 30 frames long, the transformation range would be 3 time steps; i.e. the first FCN transforms 1f-GEI to 3f-GEI; the second one transforms 3f-GEI to 6f-GEI, and so on ($i$f-GEI denotes a GEI generated from $i$ frames).
  
\begin{figure}[!h]
    \centering
    \includegraphics[width=0.99\linewidth, height=8.2cm]{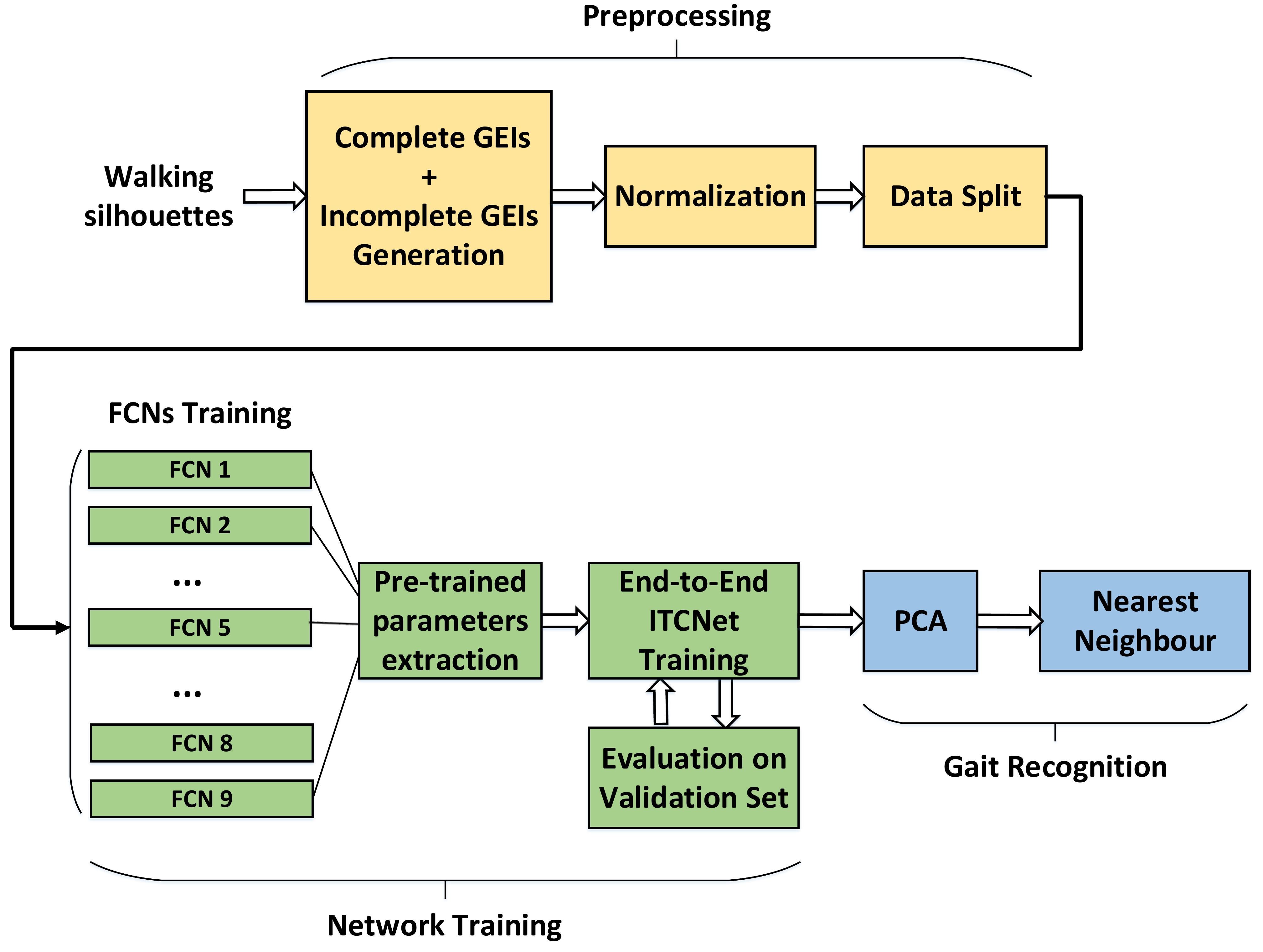}
    \caption[Illustration of the complete workflow]{Illustration of the complete workflow}
\label{fig:workflow}
\end{figure}

The complete workflow of the proposed approach is illustrated in Fig.~\ref{fig:workflow}. At first, we generate different types of incomplete GEIs as training data based on the number of frames and the starting point. These input samples are resized and normalized before being fed into the corresponding FCN. Next, we fine tune the end-to-end network on any type of incomplete GEI. Finally, the reconstructed GEIs are assessed in a recognition task. 

\subsection{Gait Energy Image}
\begin{figure}[h]
	\centering
	\includegraphics[width=0.98\linewidth, height=3.0cm]{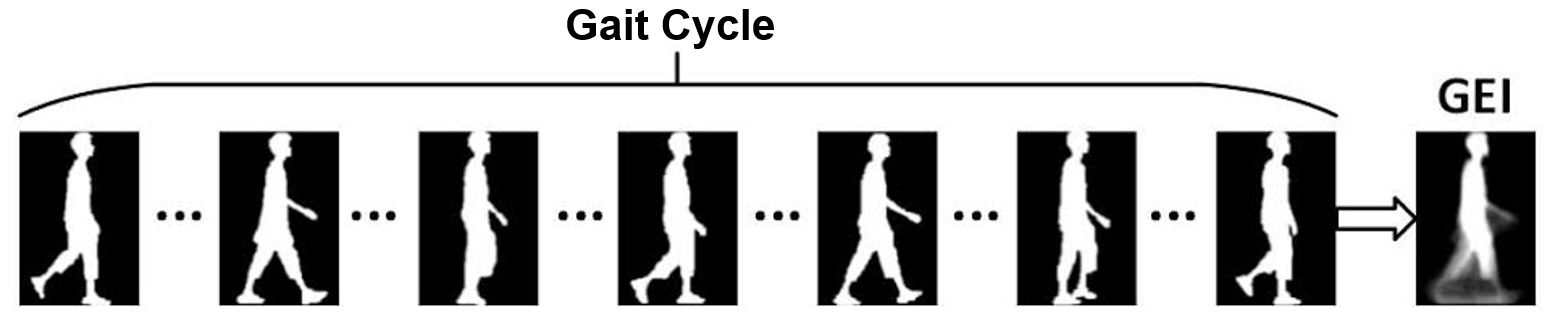}
	\caption{GEI as a gait feature is computed by averaging gait silhouettes over one gait cycle.}
	\label{fig:GEI}
\end{figure}

The gait energy image (GEI) is obtained by averaging the binary silhouette of human body over one gait cycle (see Fig.~\ref{fig:GEI}). Let $f(x, y)$ denotes the binary value of a pixel in position $(x, y)$ at time $t$, and $N$ is the number of frames in one gait cycle. The gray-level gait energy image is computed as
\begin{equation}
GEI\left (x,y  \right )= \frac{1}{N} \sum_{i=1}^{N} f_{i}\left ( x,y \right ).
\end{equation}
A pixel value in GEI shows the probability of the pixel position has been occluded by a human body during a complete gait cycle. Being an effective feature in gait recognition, GEI is used as the input and the target image type in our method. However, the proposed method could be applied to other gait features like GEnI, or EGEI. 

\subsection{Structure of FCNs}
Having mentioned before, we propose an incremental GEI reconstruction approach using 9 FCNs that each single FCN enhances the quality of input GEI. Since the gait cycle length depends on the frame rate and is different from a dataset to another dataset, we consider the partial transformations every $1/10$ of the gait cycle length. The first FCN transforms a GEI generated from the first $1/10$ of the gait cycle to the GEI corresponds to the first $2/10$ of the gait cycle. Similarly, the other FCNs enhance their incomplete GEI by predicting the information of the following $1/10$ gait cycle.

\begin{figure*}[!th]
\centering
		\includegraphics[width=0.99\linewidth, height = 4.0cm]{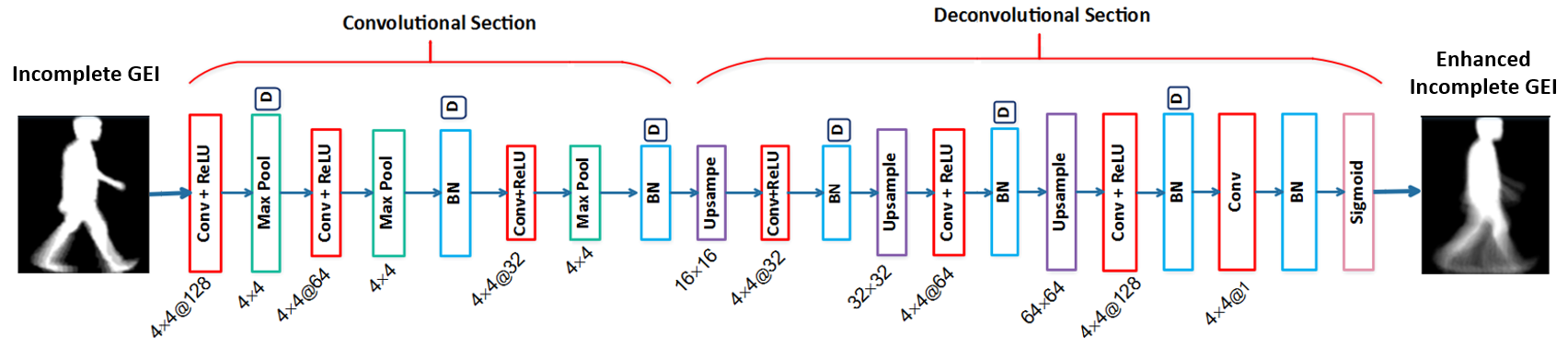}  
		\caption{The structure of the FCN for enhancing an incomplete GEI. Each rectangle shows a single computing stage. Different types of layers are displayed in different colors; Convolution and ReLU activation is in red; max pooling, green; up-sampling, purple; batch normalization, blue. The small squares noted as 'D' show dropout.}
		\label{fig:FCNN}
\end{figure*}

The structures of the all FCNs are the same, but they are trained on different types of GEI. The architecture of one FCN is shown in Fig.~\ref{fig:FCNN}. Each FCN consists of two parts; the encoder (convolutional) part and the decoder (deconvolutional) part. The encoder part contains three convolutional layers. Each convolutional layer is followed by a Relu function, which serves as a nonlinear activation function, and a pooling layer that downsamples the feature maps. After these layers, batch normalization and dropout techniques are used to accelerate convergence and avoid over fitting. 

The deconvolutional part consists of the upsampling layer, the convolutional layer, the Relu layer, batch normalization and dropout. The reason for not directly using a deconvolutional layer is that transposed convolutional layers (deconvolutional layers) can lead to artifacts such as checkerboard patterns in the final output. These effects are due to the overlap in the kernels, which can be avoided by setting the stride and kernel size to be equal. As an other alternative, Odena et al.~\cite{odena2016deconvolution} show that these checkerboard artifacts can be avoided by resizing the layers using the nearest neighbor or bilinear interpolation (upsampling) followed by a convolutional layer. We adopted this idea by selecting an upsampling and a convolutional layer as a deconvolutional layer in our network. 

The size of the input image is $64\times64$ pixels. The kernel size of each convolutional layer is set to $(4, 4)$, and the strides are set to $1$ in both horizontal and vertical directions. The output dimensions of the convolution layers are set to $128, 64$, and $32$ in turn. The kernel size and pooling strides for each pooling layer are both $(2, 2)$, which means a picture's height and width will be halved when it goes through the pooling layer.

In the output layer, we place a sigmoid activation function to have a gray-scale output image with pixel values in $[0, 1]$. The convolutional layer after the first upsampling layer is considered to be the hidden layer whose parameters are used in the end-to-end network (ITCNet).

\subsection{ITCNet: Incomplete to Complete GEI Convertor}
\begin{figure*}[!th]
\centering
		\includegraphics[width=0.99\linewidth, height = 5.5cm]{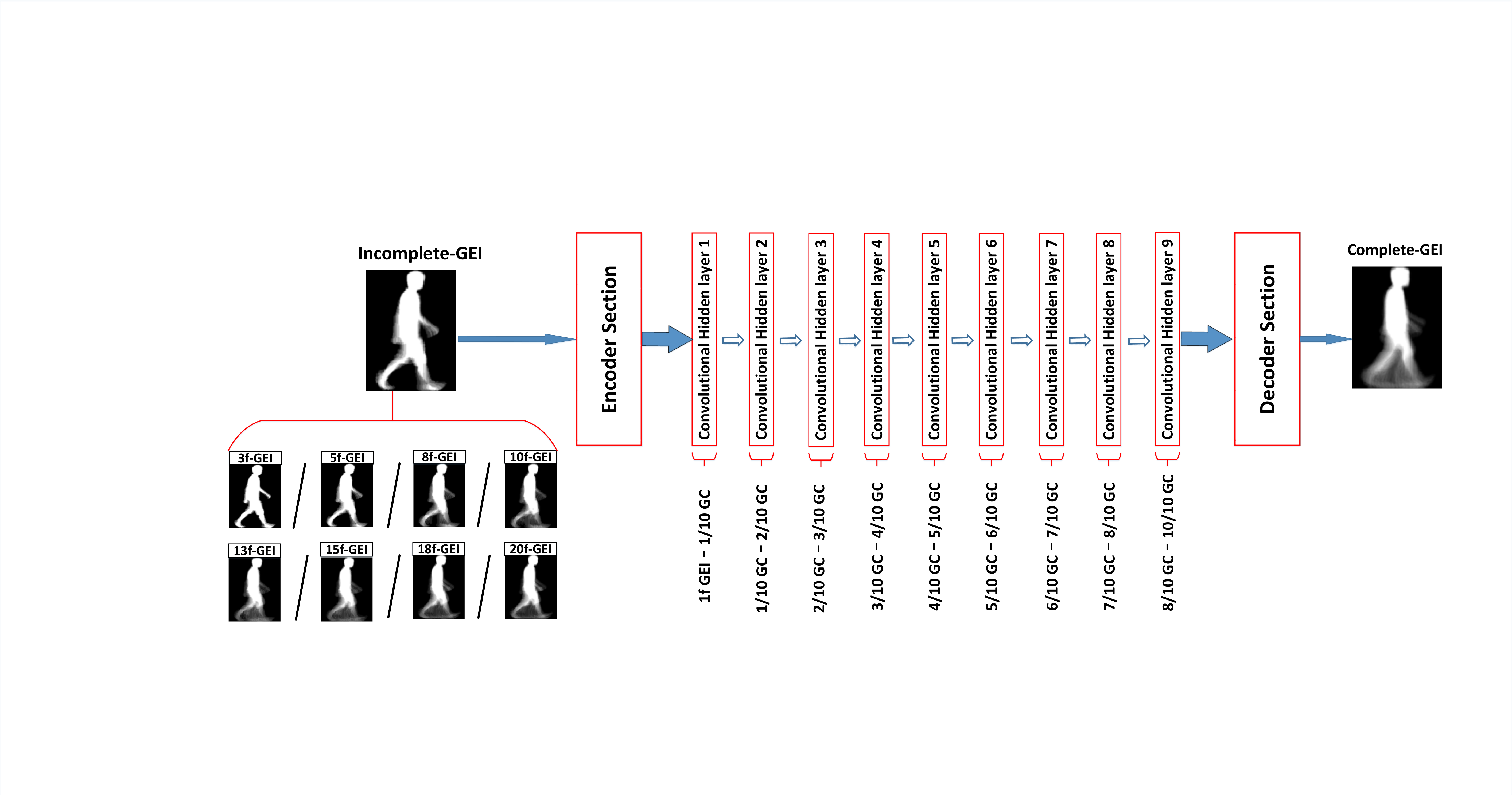}  
		\caption{The structure of the ITCNet for reconstructing a complete GEI from an incomplete GEI.}
		\label{fig:ITCNet}
\end{figure*}

In our proposed method, the complete GEI restoration is done in a progressive way (i.e. various types of incomplete GEIs are gradually converted to the complete GEI). To this end, the several trained auto-encoders are stacked together to have one end-to-end network (ITCNet). The input of the ITCNet could be any type of incomplete GEI, and the target is the corresponding complete GEI. The auto-encoder in layer $i$ would map $m$f-GEI to $n$f-GEI feature, where $n-m = T/10$, and $T$ is the gait cycle length. 
For example, if the aim is to transform a 1f-GEI to the complete GEI and the full gait cycle is $30$ frames long, the input is first mapped to 3f-GEI, then to 6f-GEI by passing through the second hidden layer, and so on.    
The structure of the ITCNet is presented in Fig.~\ref{fig:ITCNet}. All generated incomplete GEIs are used to fine tune the ITCNet. In this way, this end-to-end network can transform any type of incomplete GEI to corresponding complete GEI, without the need for any prior knowledge about the type of the input IC-GEI.

\noindent \textbf{Training:} As training data, we generate multiple incomplete GEIs from different numbers of frames in a gait cycle. Regarding the number of frames, we might have 5 or 10, or in an extreme case there might only be $1$ available. We generated different incomplete GEIs from different frame numbers $N$ and with different starting frames $M$ for training the network. All computed incomplete GEIs and their corresponding complete GEIs are resized to $64\times64$ pixels and their pixel values normalized to be in range [0, 1].

Weight initialization is an important factor for training a neural network. An improper initialization can lead to divergence in learning. Kaiming et al.~\cite{he2015delving} show the importance of the weight initialization and propose a specific initialization method for ReLU activation function. Following their findings, the weights of all convolutional layers in the presented approach are initialized using a Gaussian distribution with the standard deviation $\sigma_{c}$ from Eq.~(\ref{eq:weightInit}), where $f_{w}$, $f_{h}$, $f_{t}$ and $f_{d}$ stand for the spatial width and height, the time extent of a filter and the number of filters in a convolutional layer, respectively. All biases are initialized with zero. 
\begin{equation}
\sigma_{c} = \sqrt{\frac {2} {f_{w} * f_{h} * f_{t} * f_{d}}}.
\label{eq:weightInit}
\end{equation}

Mean Square Error (MSE) is calculated as the loss $(L)$ value between the reconstructed GEI and the corresponding complete GEI:
\begin{equation}
L = \sum\limits_{i=1}^N \|x-y\|^2
\label{eq:loss}
\end{equation}
where $x$ denotes the reconstructed complete GEI and $y$ is the original complete GEI. The loss is minimized by stochastic gradient descent algorithm using the Adam optimizer. The optimizer parameters including $\beta1, \beta2$, and $\epsilon$, are configured as $0.8$, $0.99$, and $\num{10e-8}$, respectively. The learning rate is the optimal solution through quantitative testing together with the weight decay and momentum parameter. It is initially set to $\num{e-3}$ and is decreased every $5$ epochs by a factor $10$ .
Reconstruction accuracy is computed as the portion of the errors below $0.08$ between the pixels in the ground truth image and reconstructed image. The network is trained for $50$ epochs on mini-batches of size $80$. The dropout probability is set to $0.5$ during training. 

\section{Experiments}
\label{sec:experiments} 
\subsection{Datasets} 
The proposed method was evaluated on two commonly used gait datasets, namely OULP and Casia-B.

\textbf{OULP}: The OU-ISIR Large Population (OULP) dataset is one of the most recent and popular datasets for gait recognition due to its significant amount of recorded subjects. This dataset provides gait silhouettes of $4007$ subjects in size $128\times88$ pixels, including $2135$ males and $1872$ females with ages ranging from $1$ to $94$ years old. Each subject in the OULP dataset is recorded under four different observation angles: $55^{\circ}, 65^{\circ}, 75^{\circ}, 85^{\circ}$. 
The OULP dataset contains two sections A and B. Section A includes a set of two sequences (gallery and probe sequences) of subjects and is suitable for evaluating the gait recognition performance under constant normal walking conditions, and section B includes a set of subjects with only one sequence and is used for gender-based gait classification. 

\begin{figure*} [!h]
\centering
\includegraphics[width=0.99\linewidth, height = 6.4cm]{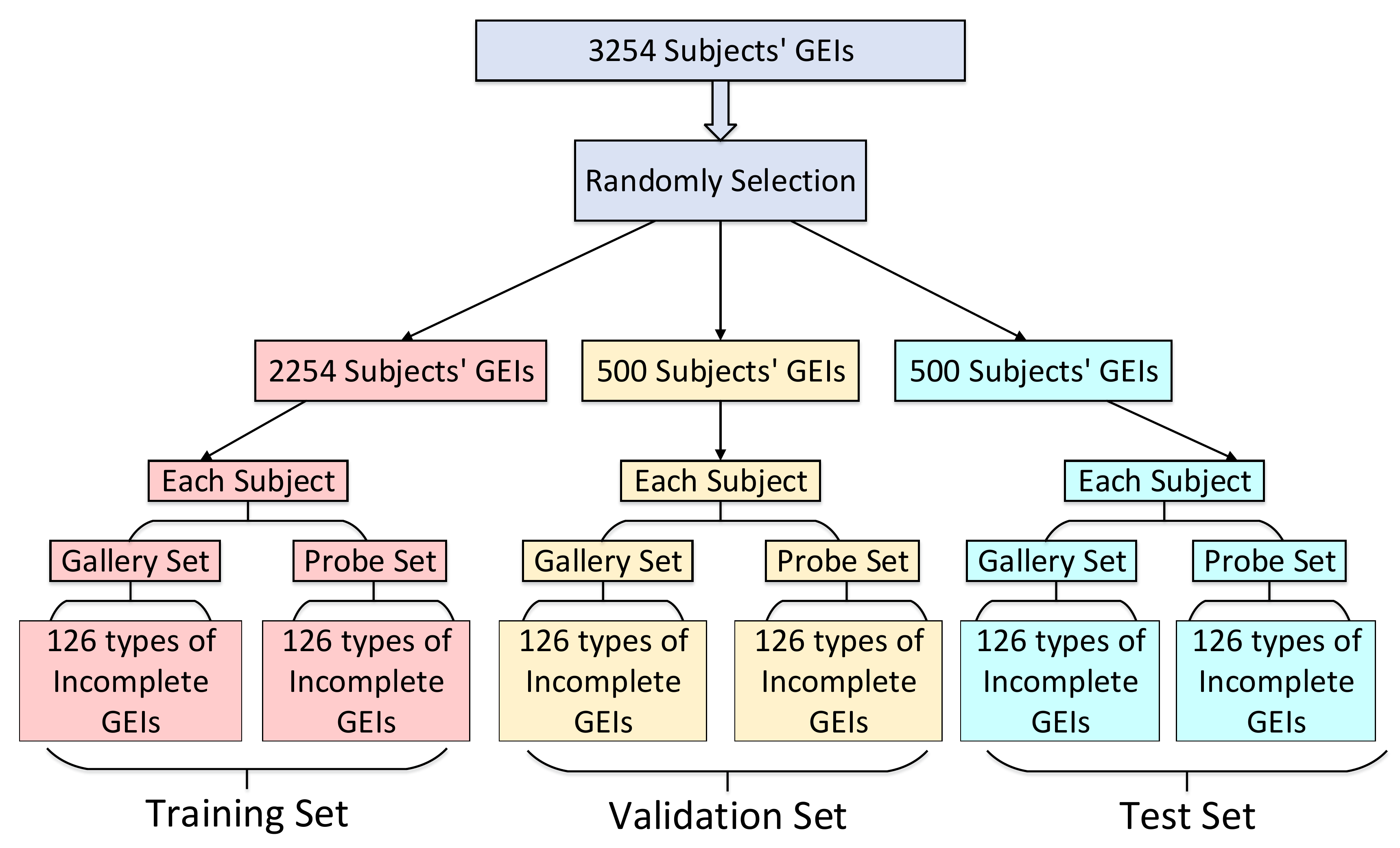}
\caption{Data splitting of the OULP dataset.}
\label{fig:data-split}
\end{figure*}

From the OULP dataset, we used the data of the section A and selected the gait data at $85^{\circ}$ to train our network, because our method does not consider the changes of gait view.  As Fig.~\ref{fig:data-split} shows, the total number of subjects in section A is $3254$; each subject has one gallery sequence and one probe sequence. Since the length of a gait cycle is 30 frames, we generated $9\times14 = 126$ IC-GEIs from each sequence ($N = 9$ and $M = 14$). For each type of IC-GEI, the corresponding TC-GEI is the same, despite the number of frames it is composed of and the starting point. That is, there is a single TC-GEI for the all IC-GEIs generated for a specific subject. As for the data split, we randomly selected $2254$ subjects from $3254$ subjects and used them as the training set, and the the remaining $1000$ subjects were put into the validation and the test sets. The inputs of the end-to-end ITCNet are all IC-GEIs, and the targets are the corresponding TC-GEIs. For each FCN, there are $14$ types of IC-GEIs for each subject. For example, to train the first FCN using OULP dataset, we only use all types of 1f-GEIs as inputs and 3f-GEIs as targets, while the uniform ITCNet uses all types of 1f-, 3f-, ..., 18f-, 20f-GEIs as inputs and corresponding TC-GEIs as targets. In short, for each FCN, $63112, 14000$, and $14000$ subjects were used for the training, validation, and test sets, respectively.

\textbf{Casia-B}: The Institute of Automation from the Chinese Academy of Sciences (CASIA) provided three datasets for gait recognition. The Casia-B dataset is one of the most widely used datasets in recent gait recognition approaches, especially for view-invariant recognition methods, because the data in Casia-B has 11 different viewing angles $(0^{\circ}, 18^{\circ}, ..., 180^{\circ})$ and different walking scenarios (normal walking, wearing coats or carrying bags). The large number of subjects and the various views of gait data in dataset B make the gait recognition tasks challenging. As we mentioned before, our method only considers TC-GEI prediction under normal walking conditions, so we only use the normal walking section of Casia-B under $90^{\circ}$ to do the experiments. The normal walking gait sequence includes $124$ subjects, with one gallery set and five probe sets for each subject. We use them all in the evaluation of network performance and gait recognition. 

\subsection{Evaluation Metrics}
In order to evaluate the quality of computed GEI, we report the reconstruction error as well as the errors in recognition tasks ( including both verification and identification), when the computed GEIs are used as data.

\textbf{Reconstruction}: As for the reconstruction evaluation, the mean square error (MSE) value of difference between each pixel of the ground truth GEI and the reconstructed GEI is computed. We also report the difference between these two images using the structural similarity index (SSIM) metric~\cite{wang2004image}. The SSIM index is a decimal value between $-1$ and $1$. In case the two images are identical, the SSIM equals value $1$.

\textbf{Recognition}:  In recognition we perform both verification and identification. For verification, we report receiver operating characteristic (ROC) curves and corresponding equal error rates (EER), while for identification we report Rank-1/Rank-5 metrics and cumulative match curve (CMC) curves. 

\subsection{Reconstruction Results} 
The reconstruction results of the end-to-end network on the OULP and the Casia-B datasets are
listed in Table~\ref{tab:test_Acc_MSE_SSIM_OULP} and Table~\ref{tab:test_Acc_MSE_SSIM_Casia}, respectively. From Table~\ref{tab:test_Acc_MSE_SSIM_OULP}, we can see that the reconstruction results of the RC-GEIs restored using the IC-GEIs that consist of more frames are closer to the true GEIs (TC-GEI). However, this upward trend has become slower with the increase of the number of frames that compose the IC-GEIs. This is because the 18-or 20f-GEIs are very close to TC-GEIs.  
The same behavior is seen in Table~\ref{tab:test_Acc_MSE_SSIM_Casia}.
\begin{table}
\caption{Reconstruction error and accuracy ($\mu \pm \sigma$) of predicted GEIs on \textbf{OULP} test set. $N$ shows the number of frames used to build incomplete GEI.}
\vspace{0.3cm}
\centering
  \begin{tabular}{c|ccc}
  \hline
	\hline
   N   &  MSE [$\times\num{e-3}$]   &  SSIM [$\times\num{e-2}$]   & Recon-Acc [\%]  \\
  \hline
  1     &  3.4 $\pm$ 2.0  &  90.1 $\pm$ 4.8  &   89.35 $\pm$ 4.41  \\
  3     &  2.7 $\pm$ 1.4  &  92.1 $\pm$ 3.4  &   91.13 $\pm$ 3.26  \\
  5     &  2.2 $\pm$ 1.0  &  93.0 $\pm$ 2.8  &   92.36 $\pm$ 2.75  \\
	8     &  1.7 $\pm$ 0.7  &  94.5 $\pm$ 2.3  &   93.74 $\pm$ 2.43  \\
  10    &  1.4 $\pm$ 0.6  &  95.1 $\pm$ 2.1  &   94.47 $\pm$ 2.20  \\
	13    &  1.3 $\pm$ 0.5  &  95.6 $\pm$ 1.8  &   95.08 $\pm$ 1.95  \\
  15    &  1.2 $\pm$ 0.5  &  95.9 $\pm$ 1.8  &   95.46 $\pm$ 1.87  \\
	18    &  1.1 $\pm$ 0.5  &  96.1 $\pm$ 1.7  &   95.75 $\pm$ 1.80  \\
  20    &  1.0 $\pm$ 0.4  &  96.5 $\pm$ 1.6  &   96.15 $\pm$ 1.72  \\
  \hline
  \end{tabular}
\label{tab:test_Acc_MSE_SSIM_OULP}
\end{table}

\begin{figure*} [!h]
\centering
\includegraphics[width=0.99\linewidth, height = 7.4cm]{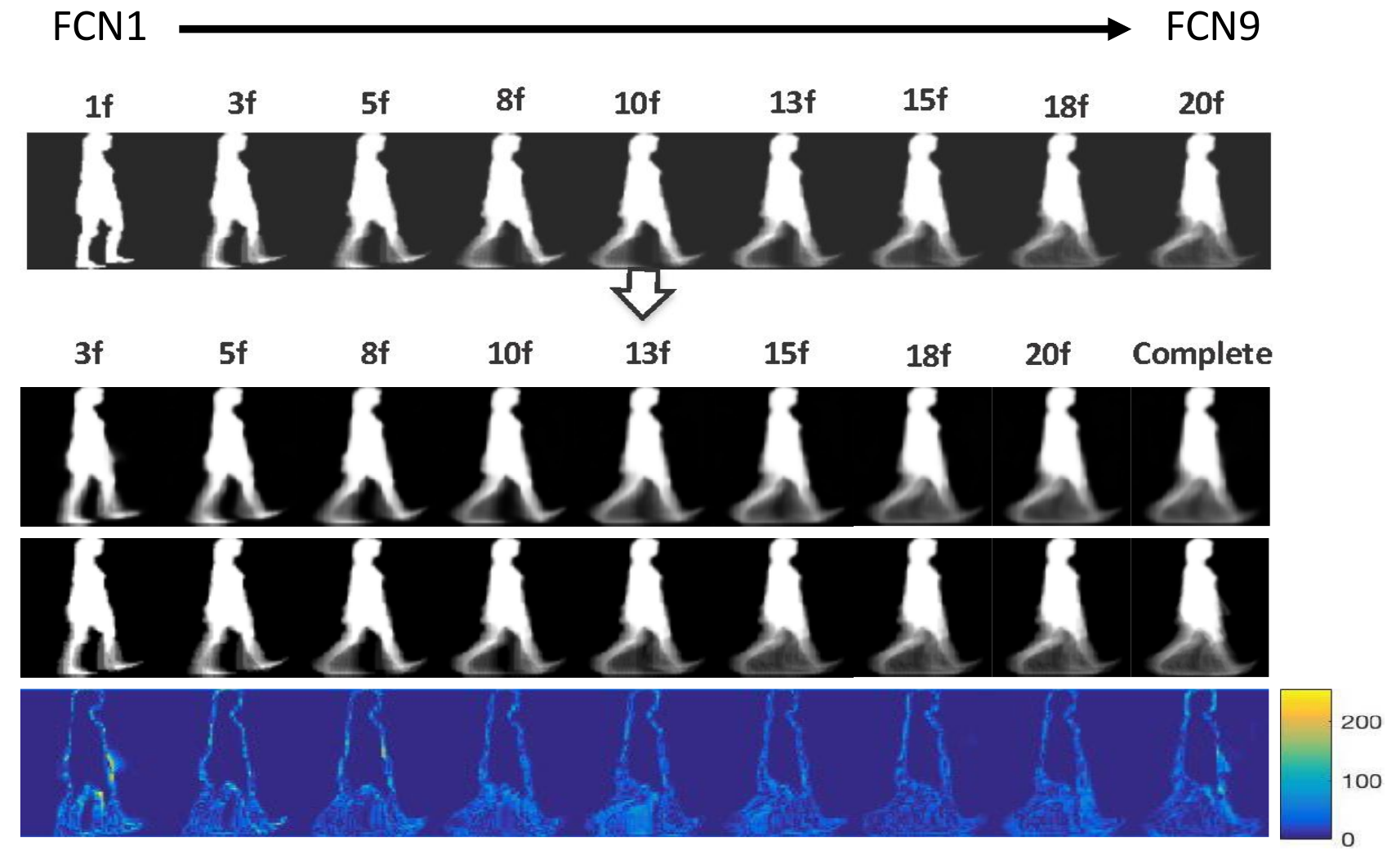}
\caption{Qualitative results of GEI reconstruction of the same subject from test set using the ITCNet.}
\label{fig:1f-20f-full}
\end{figure*}

\begin{table}
\caption{Reconstruction error and accuracy ($\mu \pm \sigma$) of test predicted GEIs on \textbf{Casia-B}. N shows the number of frames used to build incomplete GEI.}
\vspace{0.3cm}
\centering
  \begin{tabular}{c|ccc}
  \hline
	\hline
N & MSE [$\times\num{e-3}$]  &  SSIM [$\times\num{e-2}$]  & Recon-Acc [\%]\\
  \hline	
1   & 2.7 $\pm$ 0.9 & 93.5 $\pm$ 1.7 & 92.27 $\pm$ 1.89\\
2   & 2.3 $\pm$ 0.8 & 94.0 $\pm$ 1.8 & 92.99 $\pm$ 1.88\\
4   & 2.1 $\pm$ 0.7 & 94.6 $\pm$ 1.7 & 93.71 $\pm$ 1.86\\
6   & 2.0 $\pm$ 0.7 & 95.0 $\pm$ 1.6 & 94.00 $\pm$ 1.72\\
8   & 1.9 $\pm$ 0.6 & 95.2 $\pm$ 1.6 & 94.16 $\pm$ 1.61\\
10  & 1.8 $\pm$ 0.6 & 95.3 $\pm$ 1.6 & 94.39 $\pm$ 1.67\\
13  & 1.7 $\pm$ 0.6 & 95.3 $\pm$ 1.6 & 94.43 $\pm$ 1.71\\
15  & 1.7 $\pm$ 0.6 & 95.4 $\pm$ 1.6 & 94.47 $\pm$ 1.71\\
17  & 1.7 $\pm$ 0.6 & 95.4 $\pm$ 1.6 & 94.48 $\pm$ 1.70\\
  \hline
  \end{tabular}
	\label{tab:test_Acc_MSE_SSIM_Casia}
\end{table}
Figure~\ref{fig:1f-20f-full} shows the reconstruction results of different IC-GEIs using the end-to-end ITCNet, where all of the samples belong to the same subject. The input IC-GEIs start from frame 1, but they are different in the number of frames. The first row lists the IC-GEIs composed of the different number of frames sorted from smallest to largest. The reconstructed complete GEIs (RC-GEIs) and corresponding ground truth (TC-GEI) are shown in the second and the third rows, and the last row shows the color difference maps between RC-GEIs and TC-GEIs. From this figure, we can see that the RC-GEIs recovered by our end-to-end
ITCNet are almost identical to the ground truths, and the difference between ground truth and the reconstruction is mostly small. Moreover, the RC-GEIs reconstructed from IC-GEIs composed of more frames have smaller differences with their TC-GEIs. For example, the difference map of the RC-GEIs restored from 1f-GEI is significantly brighter than that of the RC-GEIs reconstructed from 20f-GEI. This shows, as expected, that the end-to-end model outputs better results close to TC-GEI, when the input IC-GEIs are composed of more frames. 

\subsection{Recognition Results} 
To evaluate the quality of the reconstructed GEIs, we conducted recognition experiments on both OULP and Casia-B datasets. The Euclidean distance was used as the similarity metric between two GEI samples in the recognition experiments. 

\subsubsection{Gait Identification}
\begin{figure}[!h]
\centering
		\begin{tabular}{ccc}
		
      \includegraphics[height=3.3cm]{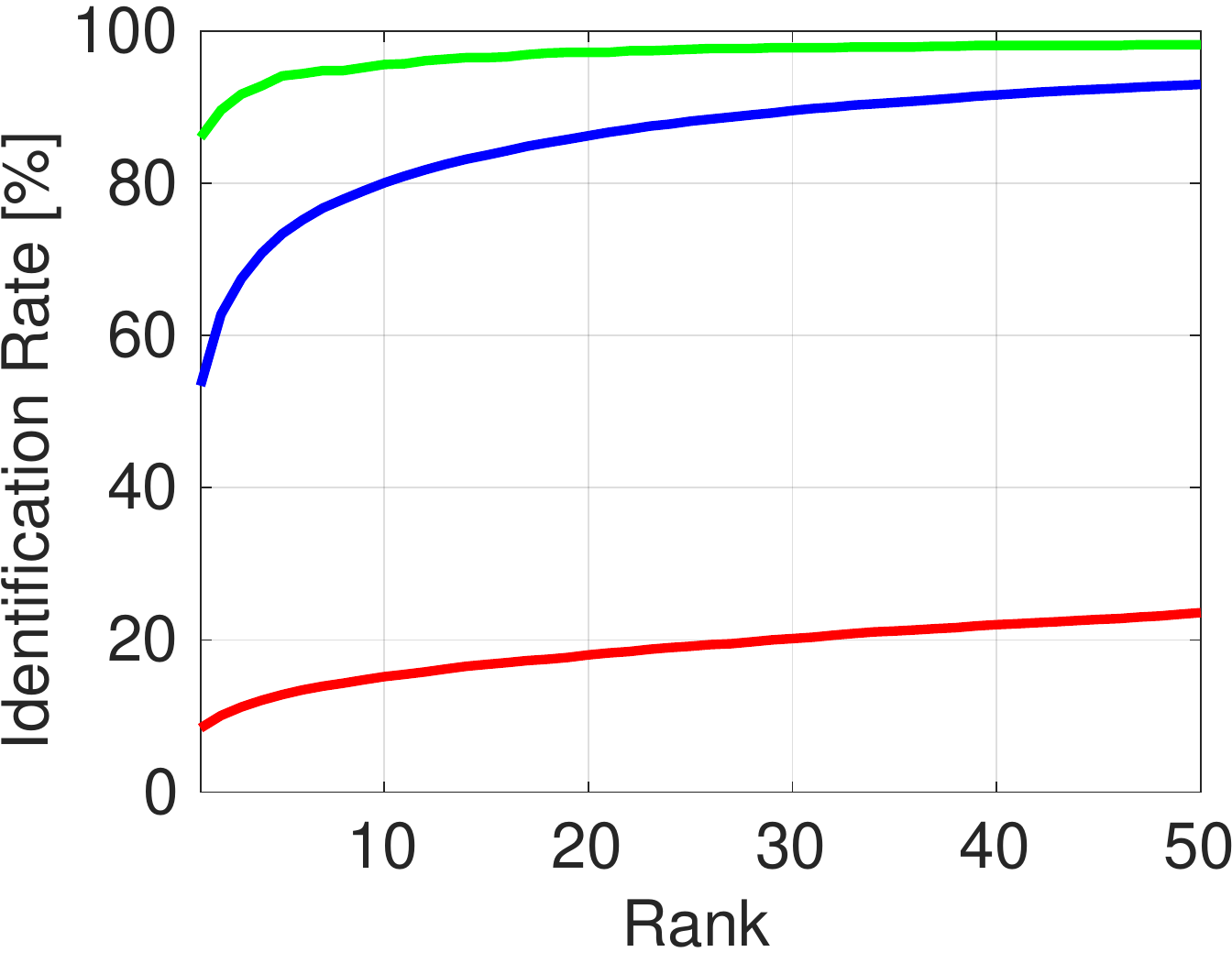} & 
			\hspace{-3mm}
    \includegraphics[height=3.3cm]{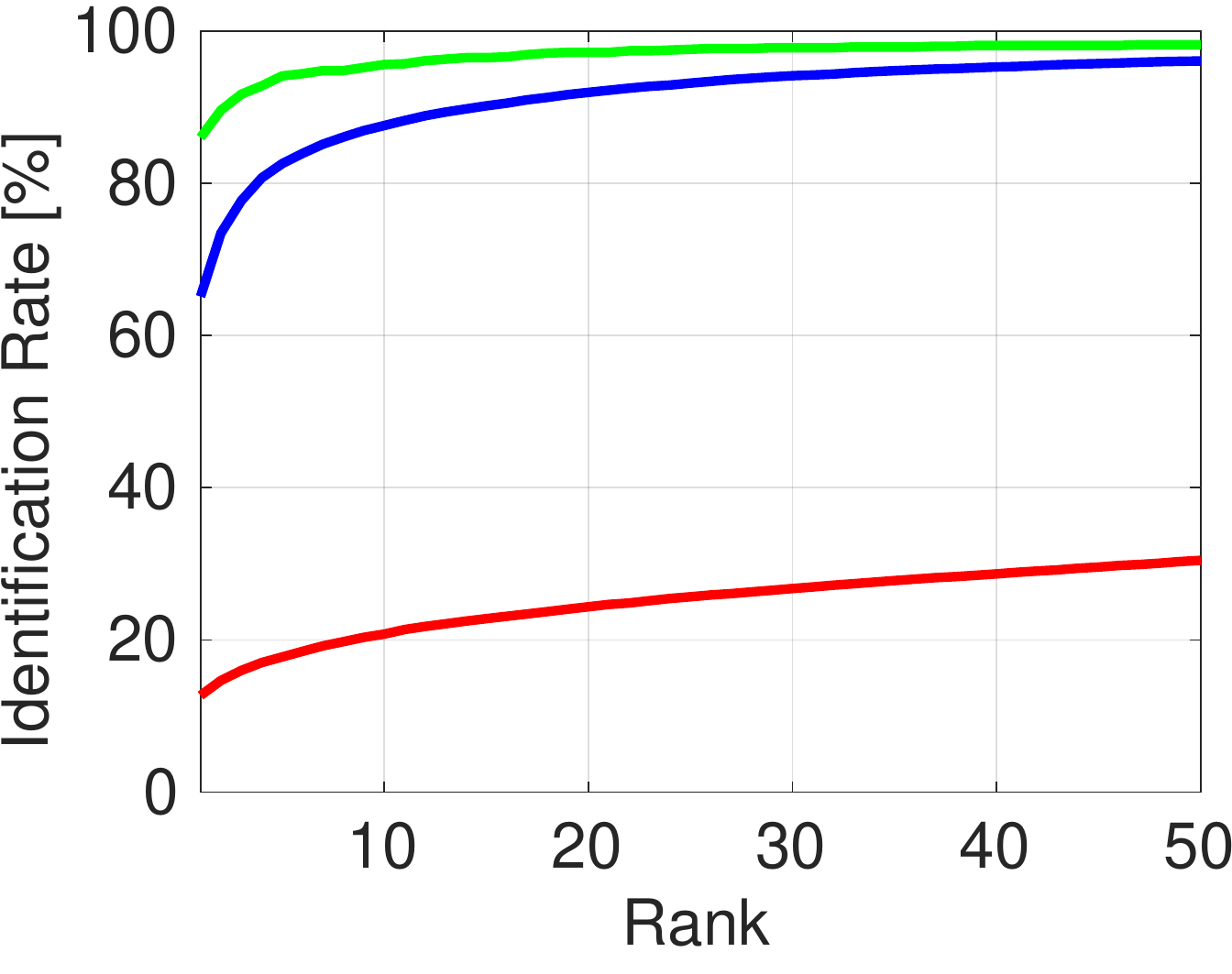} & 
		\hspace{-3mm}
		\includegraphics[height=3.3cm]{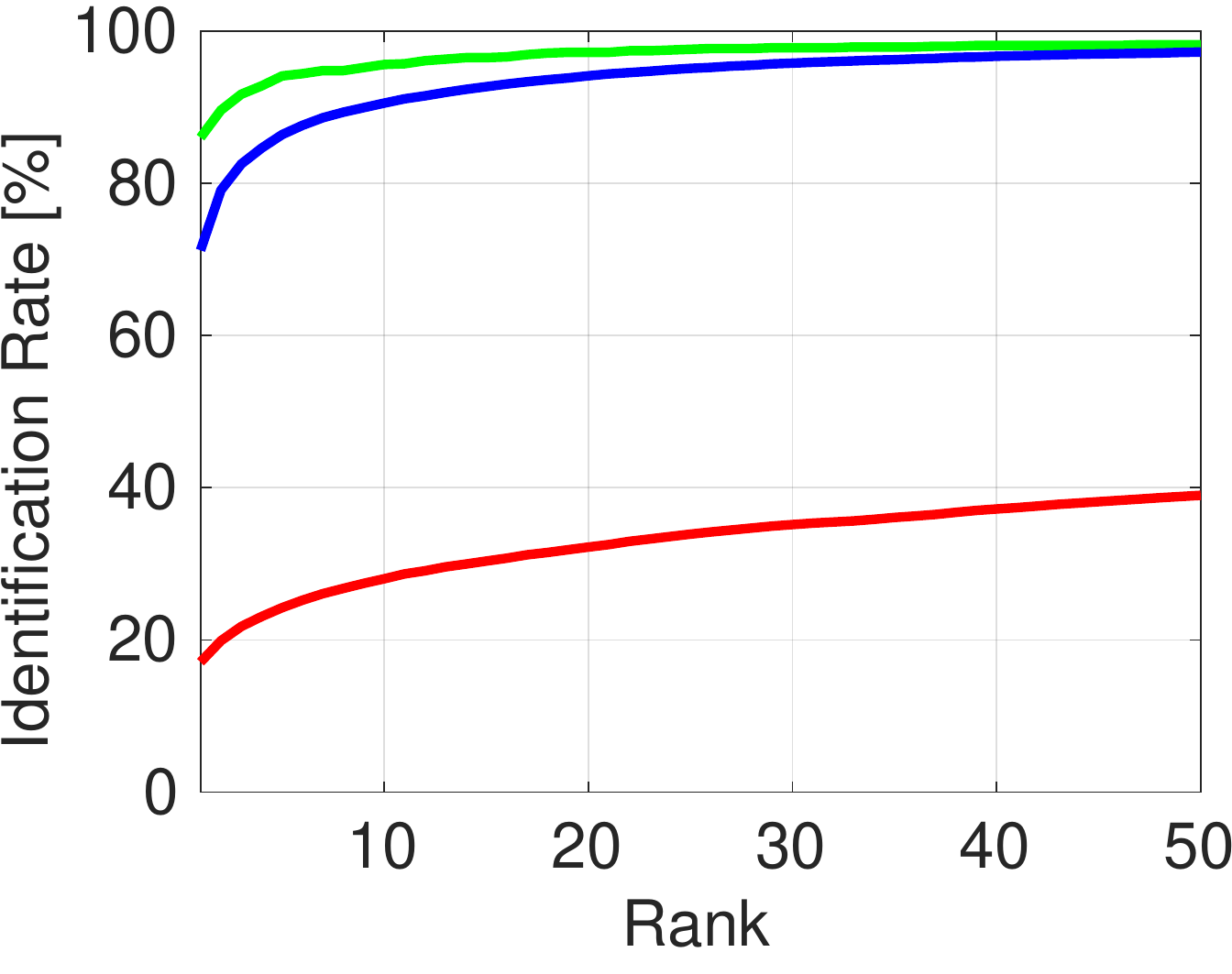} \\[-4mm]
		(a) 1 frame & (b) 3 frames & (c) 5 frames\\[4mm]
   \includegraphics[height=3.3cm]{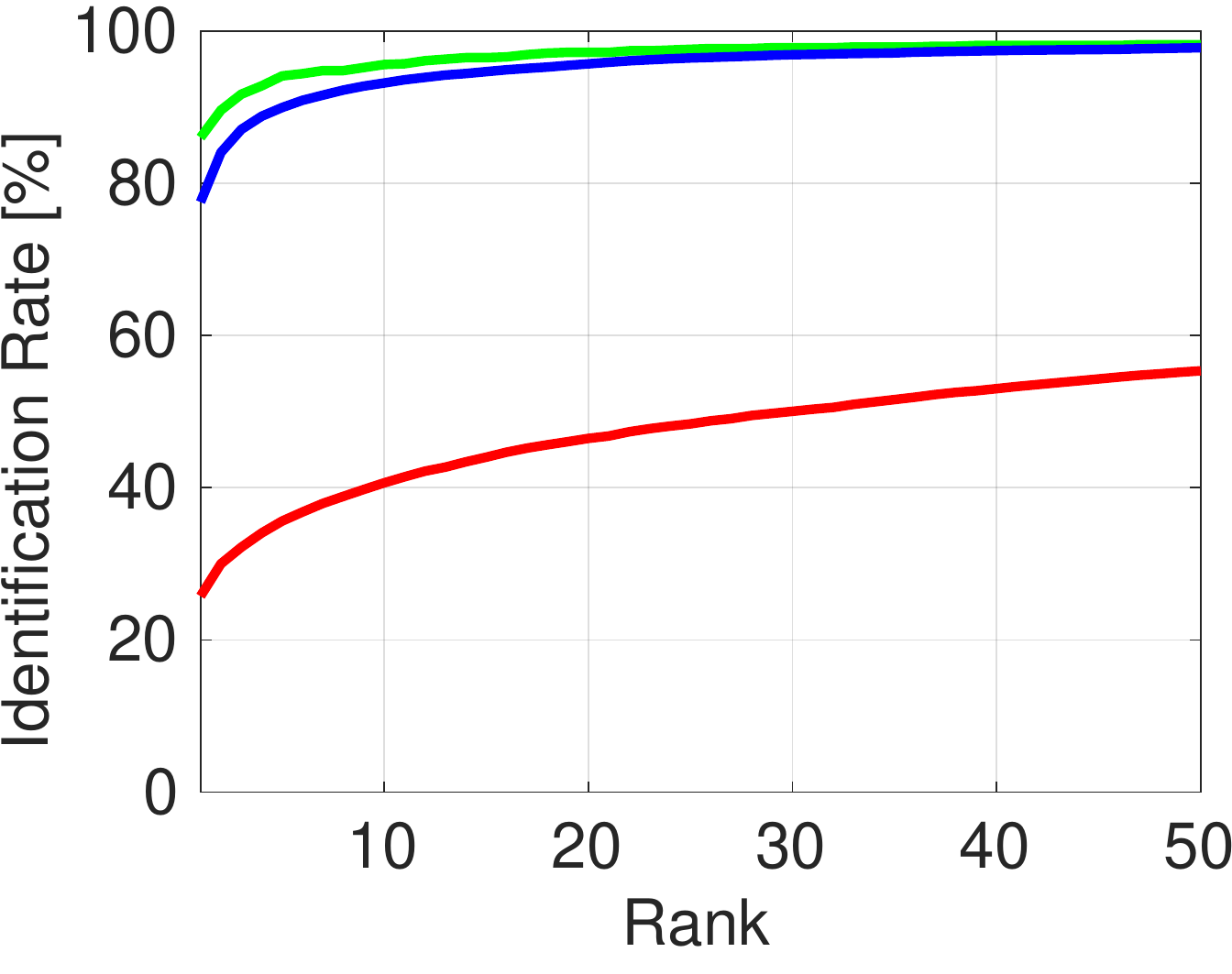} &
	\hspace{-3mm}
	 \includegraphics[height=3.3cm]{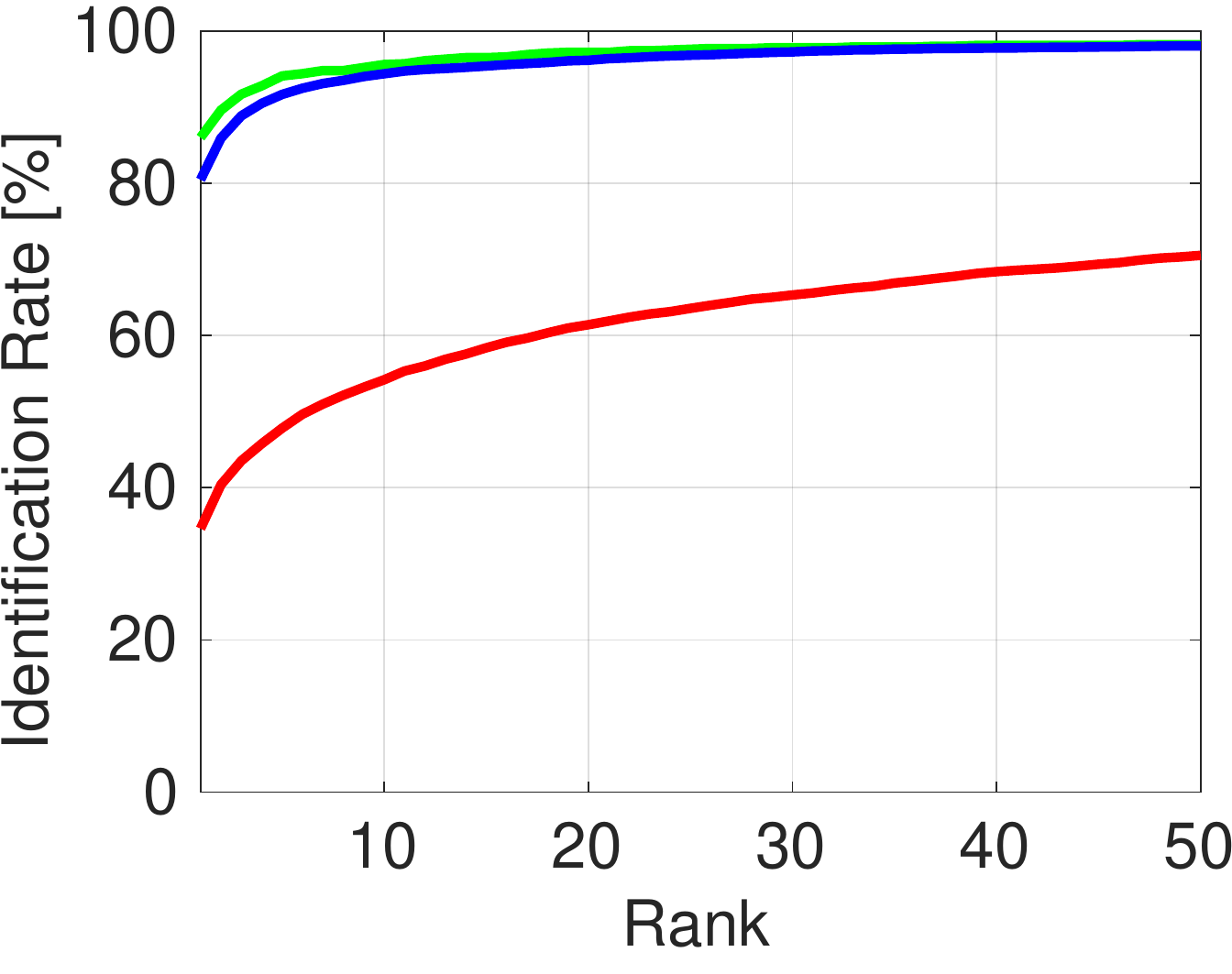} &
	\hspace{-3mm}
	\includegraphics[height=3.3cm]{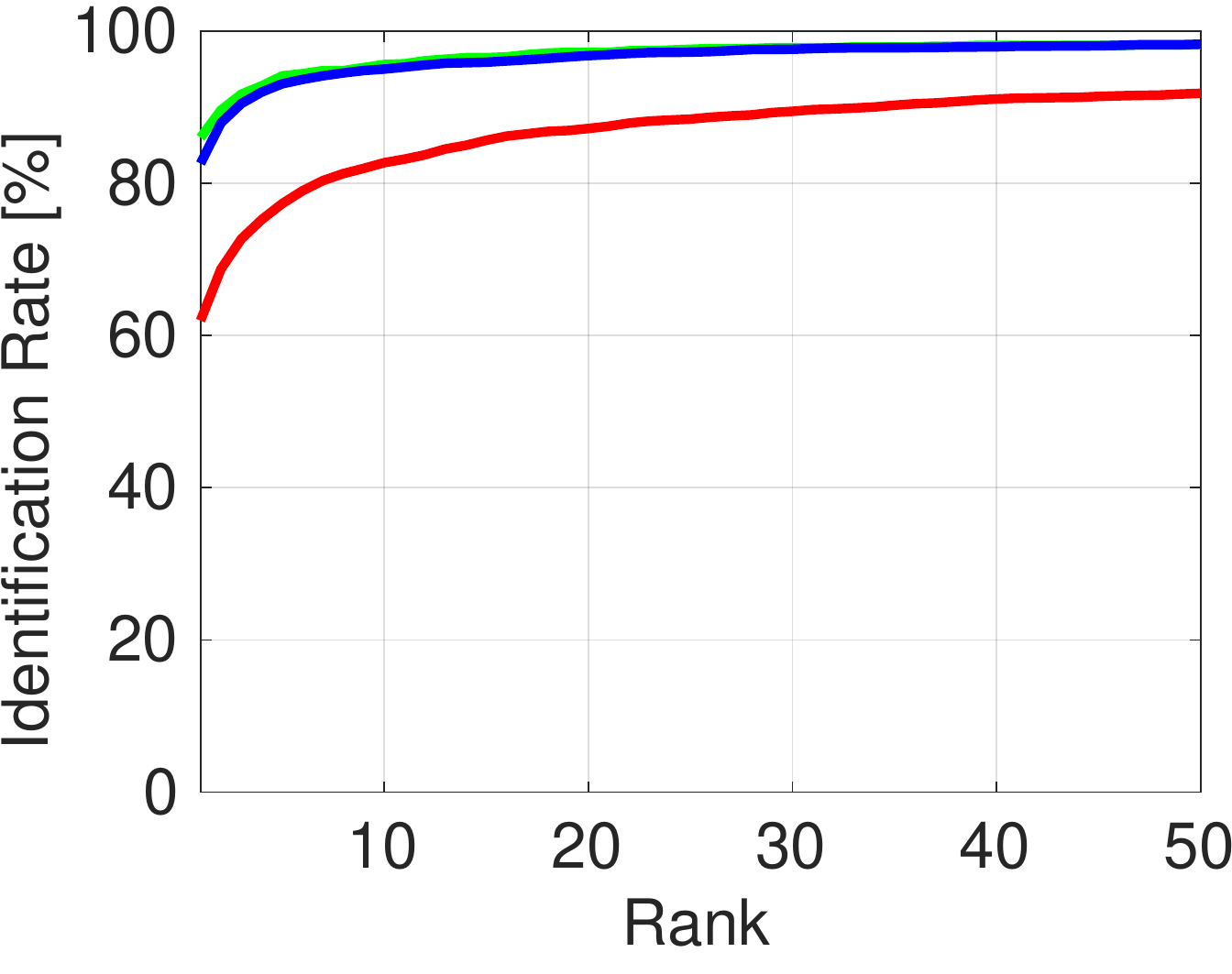} \\[-4mm]
	 (d) 8 frames & (e) 10 frames & (f) 13 frames\\[4mm]
	\includegraphics[height=3.3cm]{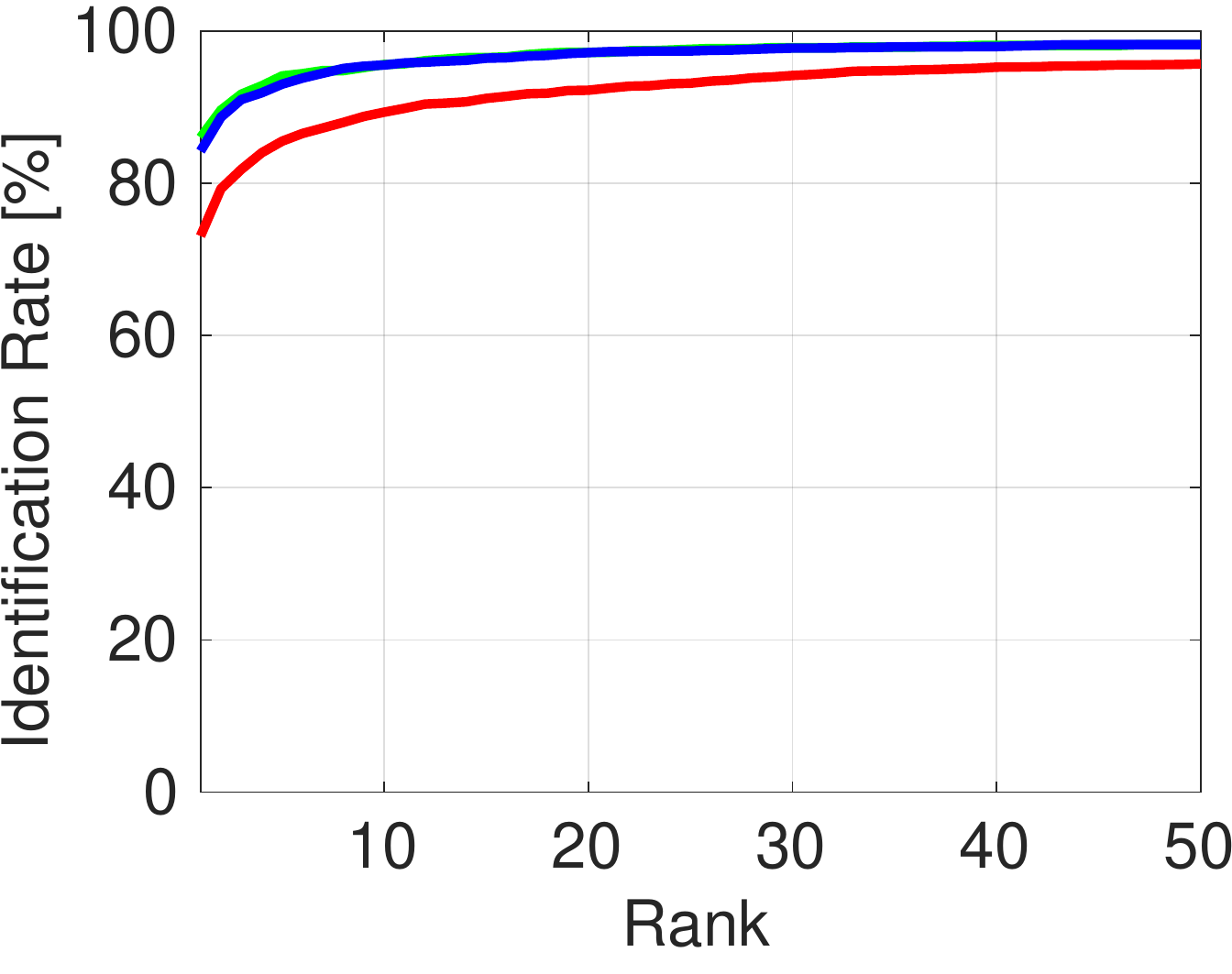}&
	\hspace{-3mm}
	\includegraphics[height=3.3cm]{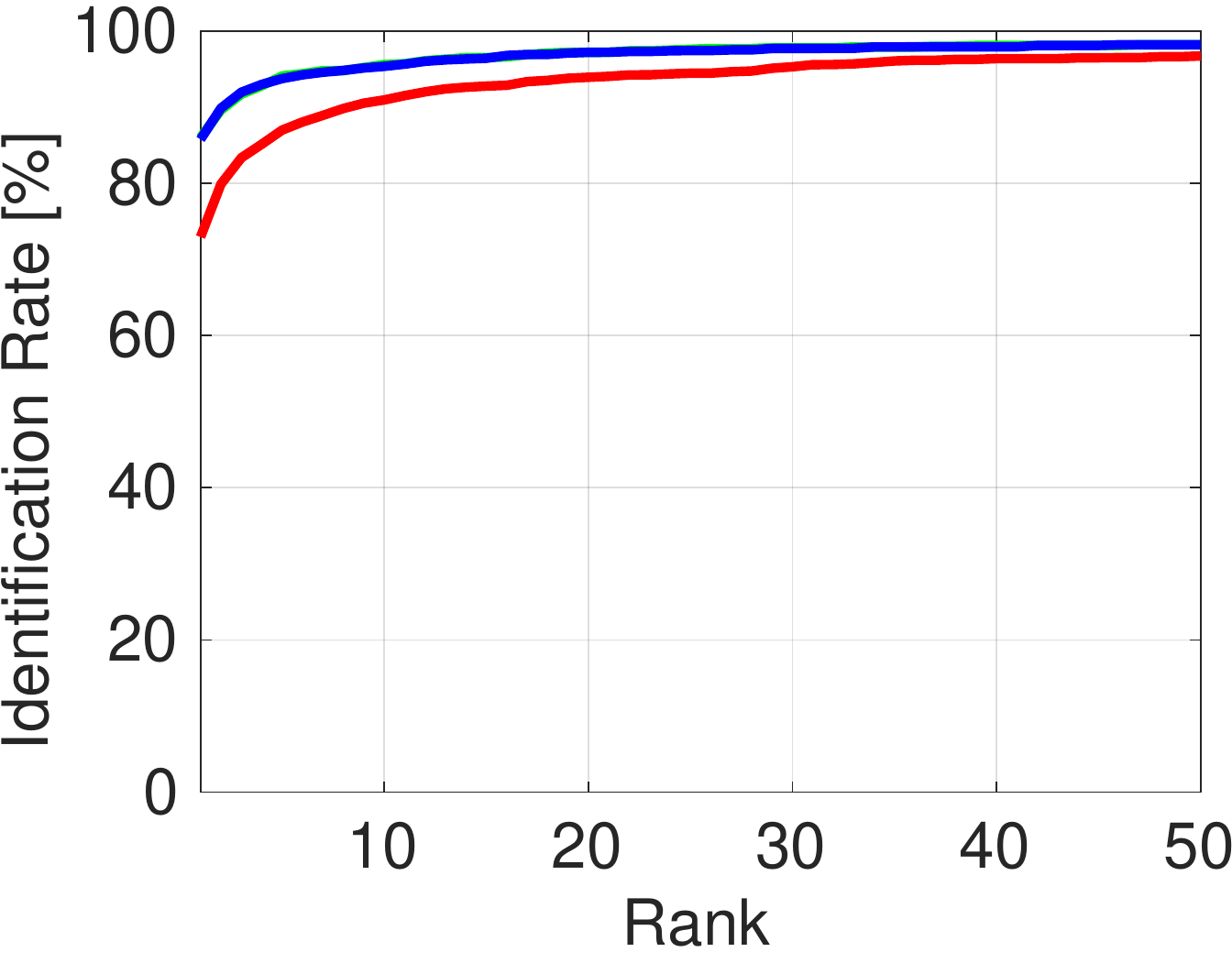} &
	\hspace{-3mm}
	\includegraphics[height=3.3cm]{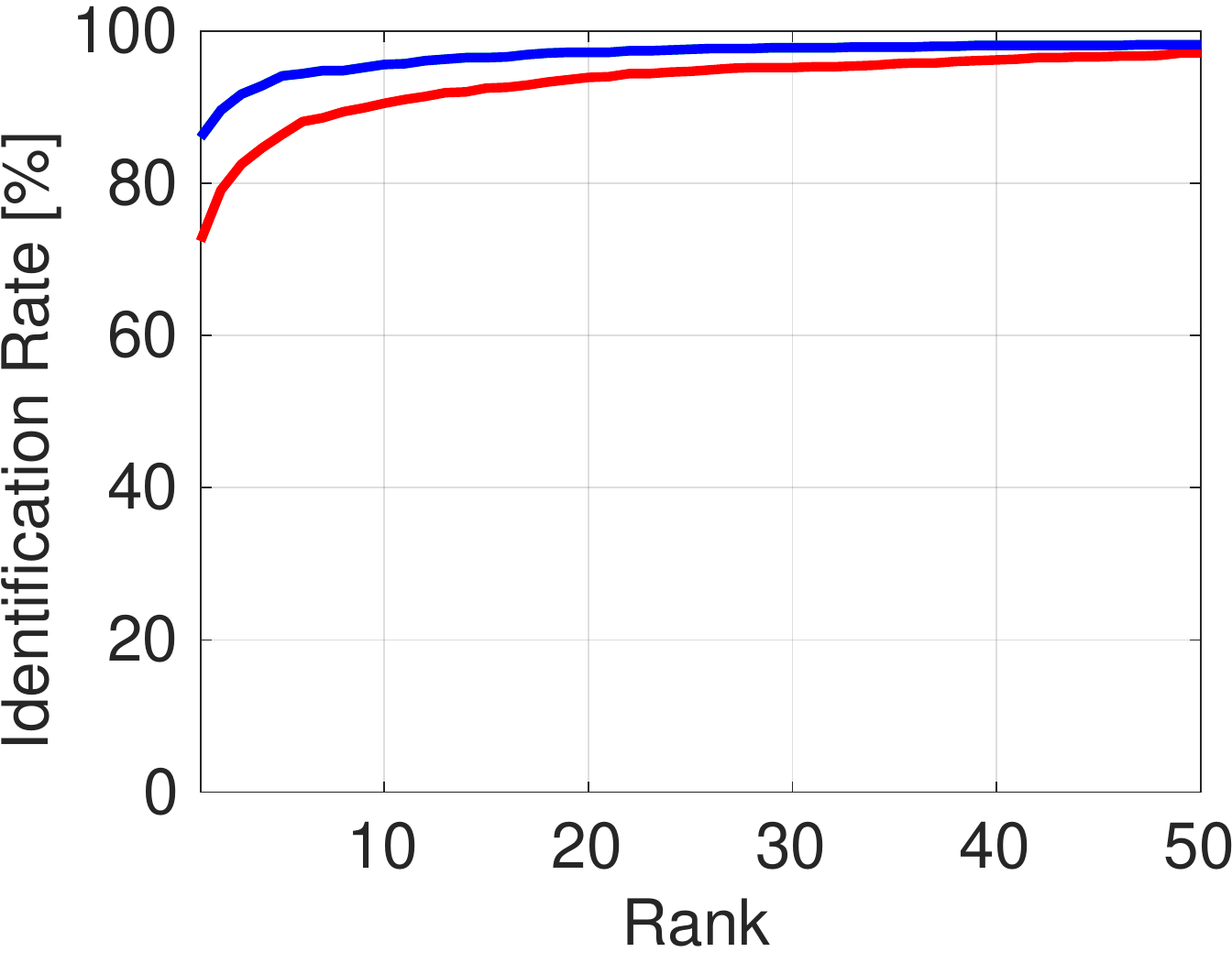} \\[-4mm] 
	(g) 15 frames & (h) 18 frames & (i) 20 frames\\
	
\end{tabular}
\caption{Identification results of the OULP test data; The CMCs of the reconstructed GEIs (blue), incomplete GEIs (red), and true GEIs (green), for different number of frames. }
\label{fig:cmc}
\end{figure}

In gait identification, a probe sample is classified as same class as its closest gallery sample based on the similarity score. The gait identification was performed on $500$ subjects of the OULP dataset. Table~\ref{tab:rank1-5_oulp} presents the rank-1 and rank-5 identification rates. Clearly, the end-to-end ITCNet has greatly improved the identification performance for IC-GEIs, especially for IC-GEIs consisting of fewer frames. As the number of frames increases, the rank-1 and rank-5 identification rates get closer to that of TC-GEIs. This trend is clearly illustrated by CMC curves in Fig.~\ref{fig:cmc}, comparing identification rates in three cases; incomplete GEI (IC-GEI), reconstructed complete GEI (RC-GEI), and true complete GEI (TC-GEI).

\begin{table}[!h]
\caption{Comparison of rank-1 and rank-5 identification rate for incomplete GEIs and reconstructed GEIs from the OULP test data. $N$ shows the number of frames used to build incomplete GEI.}
\centering
  \begin{tabular}{c|cc|cc}
  \hline
	\hline
	& \multicolumn{2}{c|}{Rank-1[\%]}  &  \multicolumn{2}{c}{Rank-5[\%]}  \\
	\hline
   N &Reconstructed GEI & Incomplete GEI & Reconstructed GEI & Incomplete GEI \\
  \hline
 1  & 53.40  &  8.42  &    73.35   & 12.81 \\
 3  & 65.11  &  12.70 &    82.57   & 17.76\\
 5  &  71.19 &  17.09 &    86.39   & 24.24 \\
 8  &  77.53 &  26.50 &    89.95   & 35.62 \\
 10 &  80.46 &  34.63 &    91.67   & 47.83 \\
 13 &  82.59 &  61.96 &    93.06   & 77.31 \\
 15 &  84.22 &  72.40 &    93.02   & 85.53 \\
 18 &  85.76 &  72.94 &    93.75   & 86.98 \\
 20 &  86.00 &  73.06 &    94.10   & 86.40 \\   
  \hline
  \end{tabular}
\label{tab:rank1-5_oulp}
 \end{table}

\begin{table}[!h]
\caption{Comparison of identification rate [\%] for incomplete GEIs and reconstructed GEIs from the Casia-B test data. $N$ shows the number of frames used to build incomplete GEI.}
\centering
  \begin{tabular}{c|c|c}
  \hline
	\hline
   N &Reconstructed GEI & Incomplete GEI  \\
  \hline
 1  &  50.01 &   30.00 \\
 2  &  50.09 &   39.05 \\
 4  &  60.02 &   40.10 \\
 6  &  75.10 &   50.04 \\
 8  &  80.00 &   50.20 \\
 10 &  80.06 &   55.30 \\
 13 &  77.12 &   70.00 \\
 15 &  85.24 &   75.00 \\
 27 &  85.30 &   76.05 \\   
  \hline
  \end{tabular}
\label{tab:rank1_casia}
 \end{table}

\subsubsection{Gait Verification}
\begin{figure}[!h]
\begin{tabular}{ccc}
    \includegraphics[height=3.3cm]{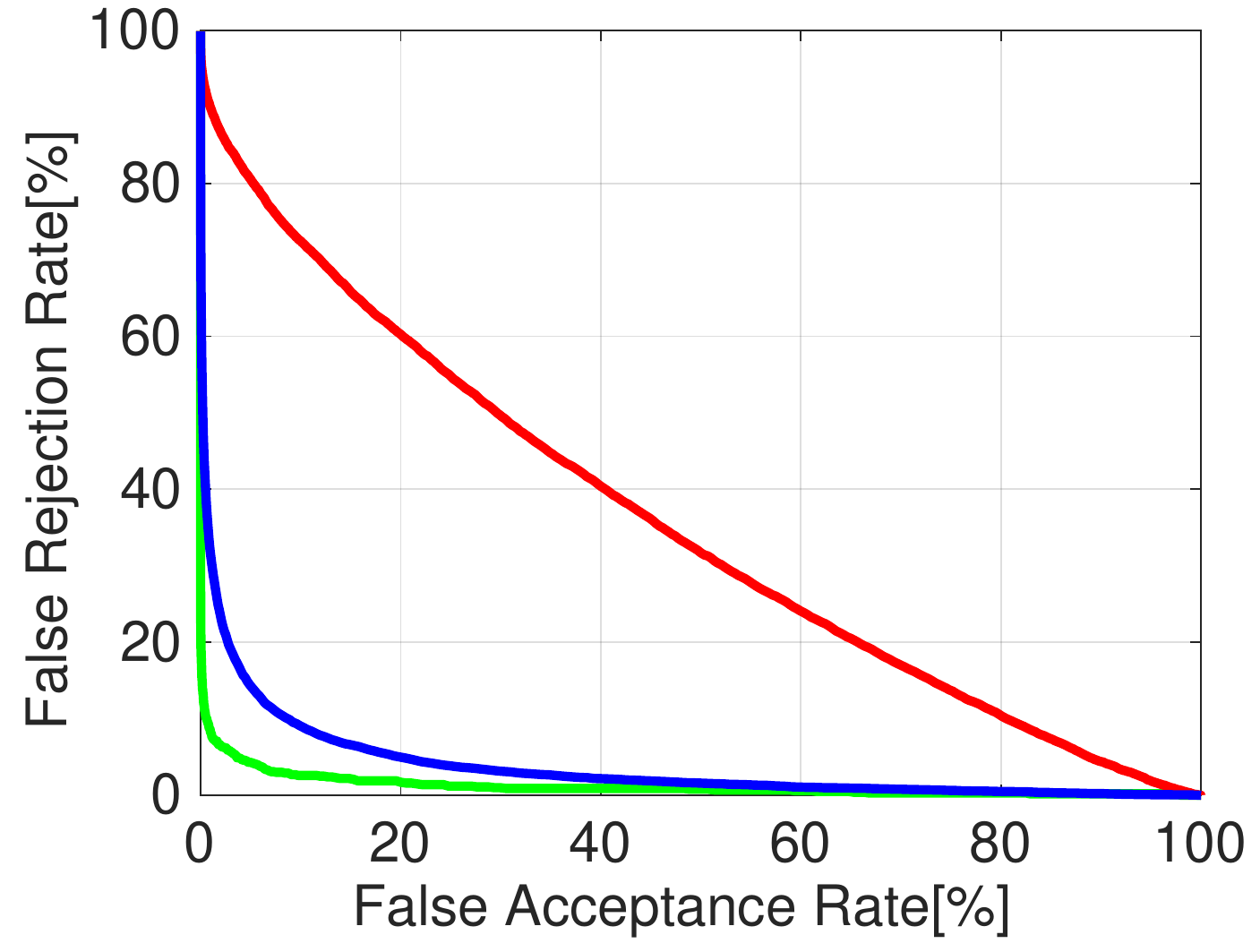} & 
		\hspace{-3mm}
    \includegraphics[height=3.3cm]{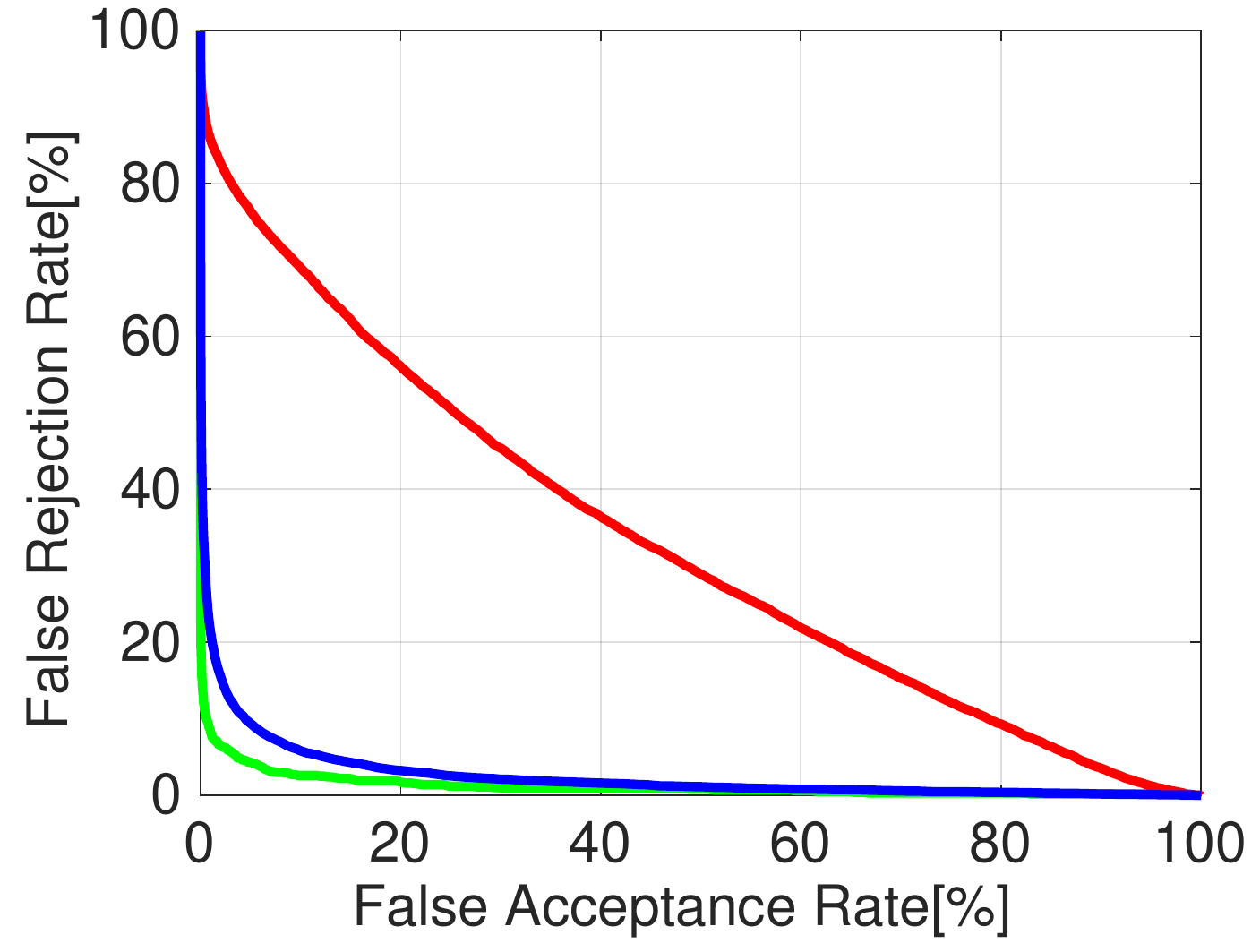} &
		\hspace{-3mm}
		\includegraphics[height=3.3cm]{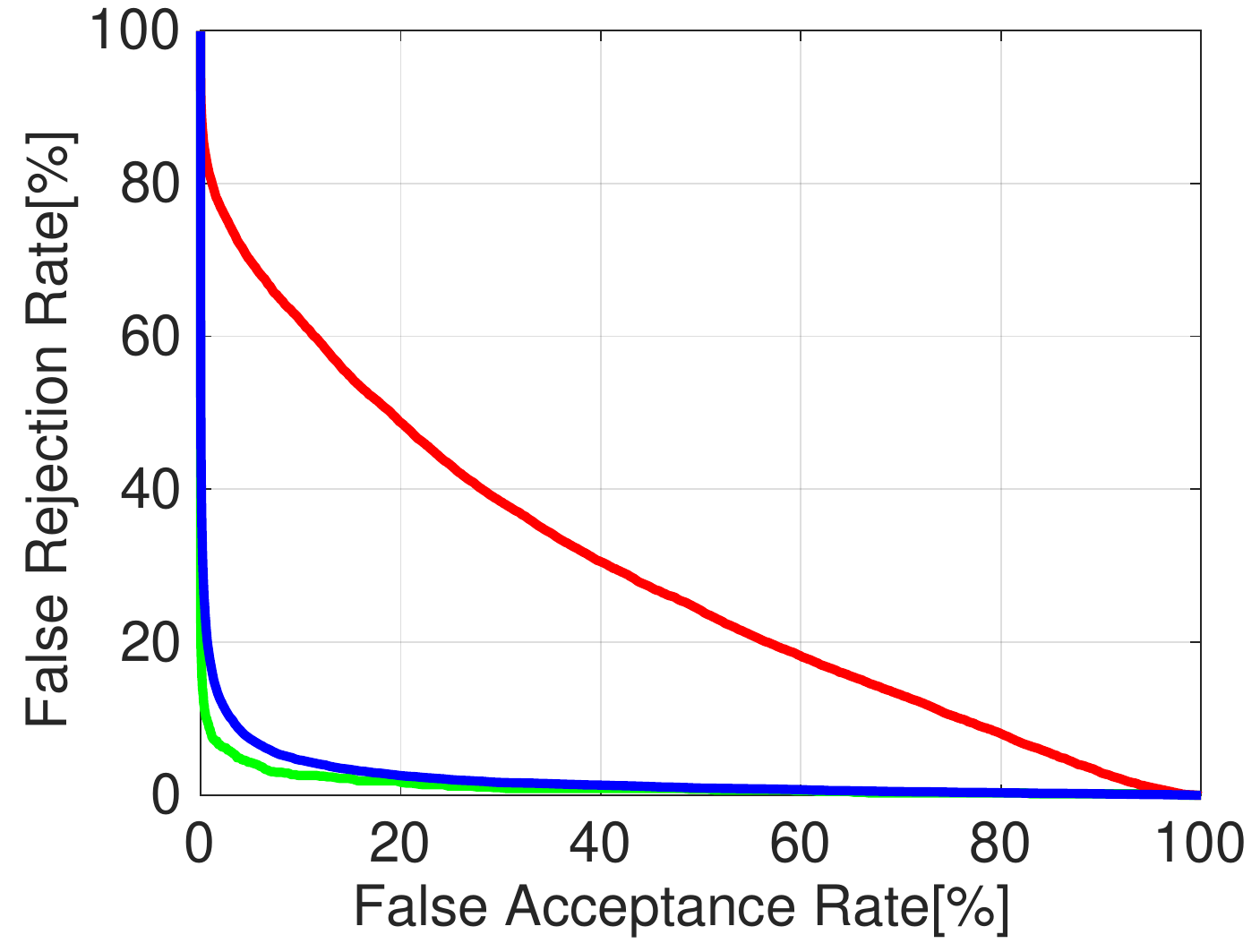} \\
    (a) 1 frame & (b) 3 frames & (c) 5 frames\\[2mm]
		\includegraphics[height=3.3cm]{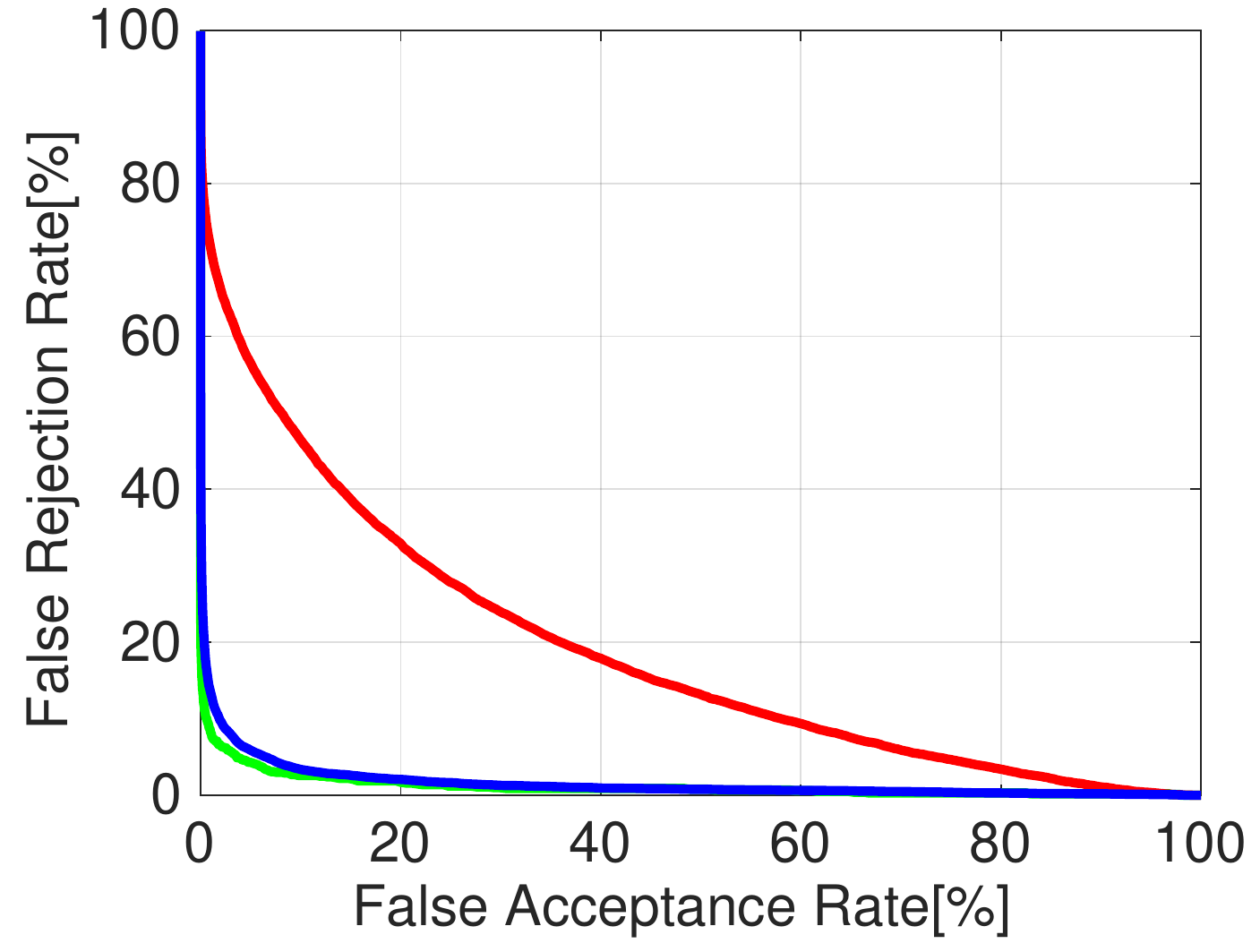} &
		\hspace{-3mm}
		\includegraphics[height=3.3cm]{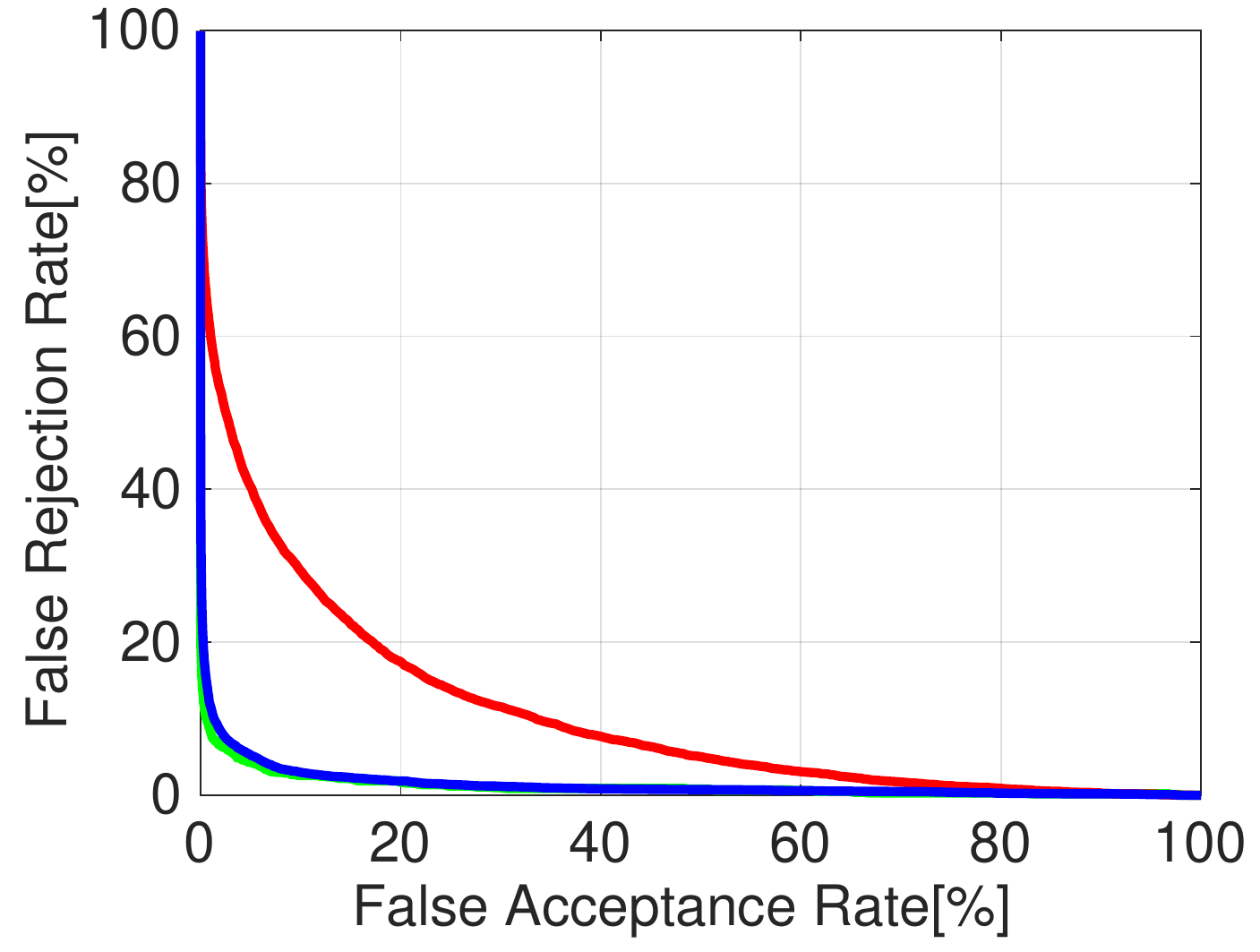} &
		\hspace{-3mm}
		\includegraphics[height=3.3cm]{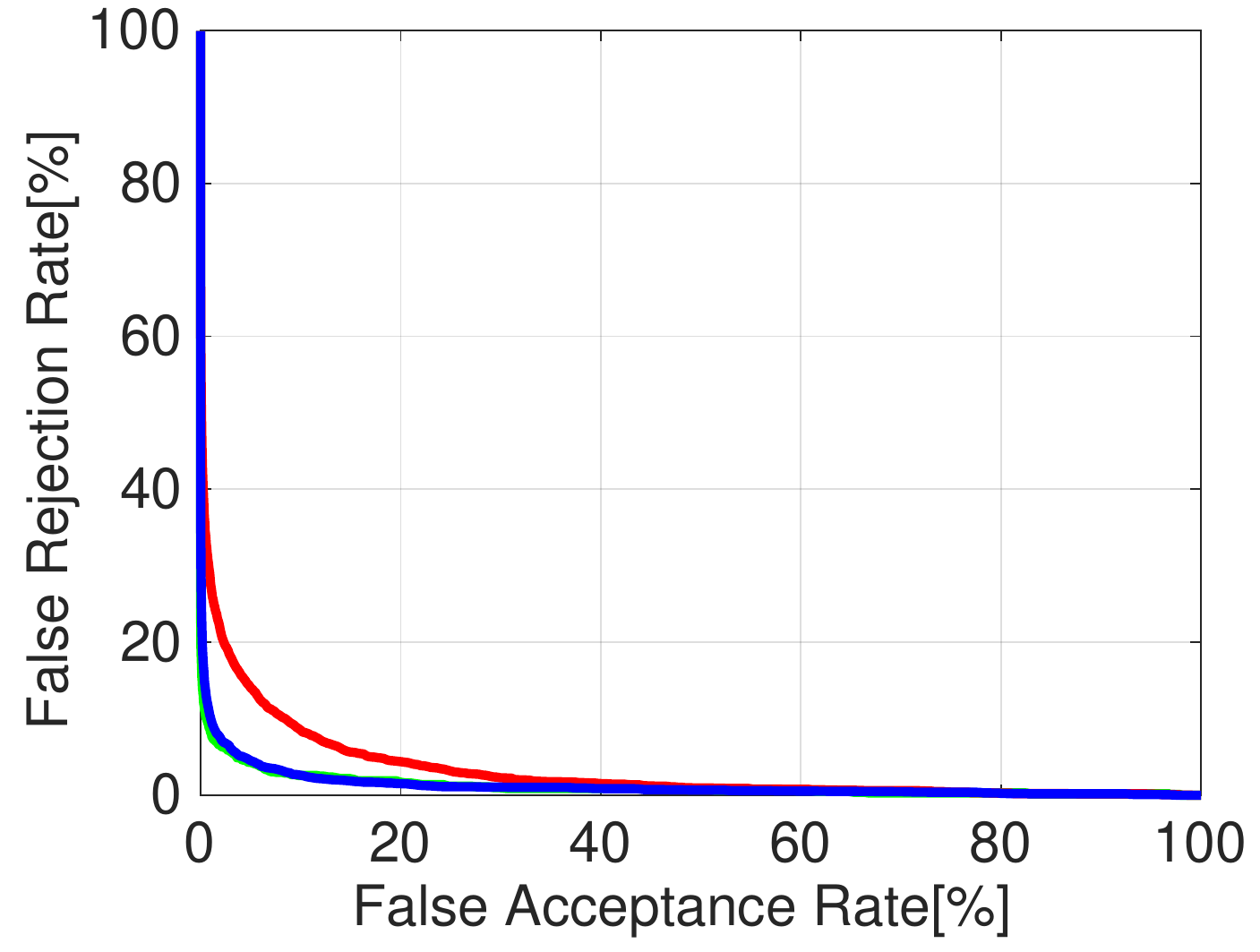} \\
		(d) 8 frames & (e) 10 frames & (f) 13 frames\\[2mm]
		
		\includegraphics[height=3.3cm]{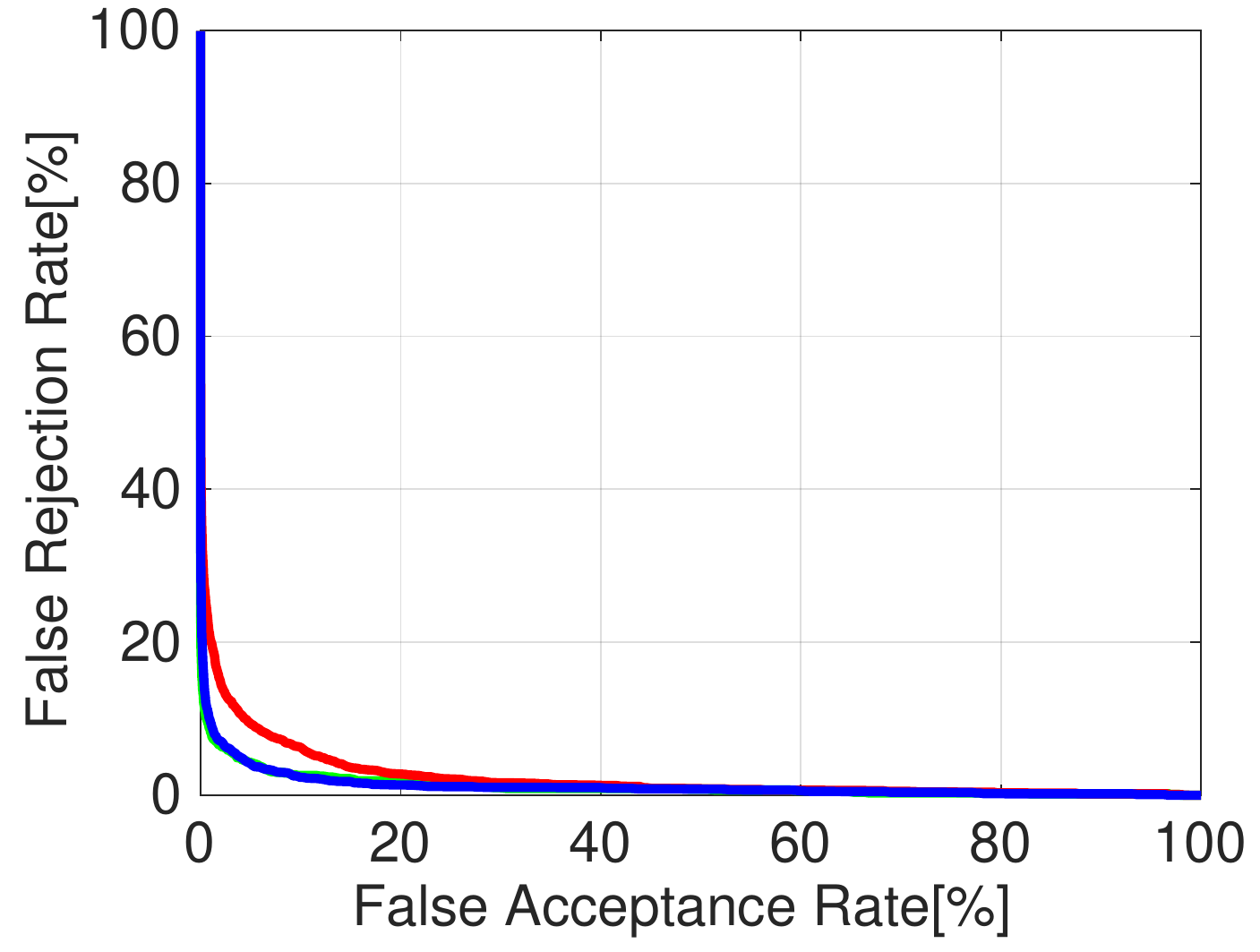} &
		\hspace{-3mm}
		\includegraphics[height=3.3cm]{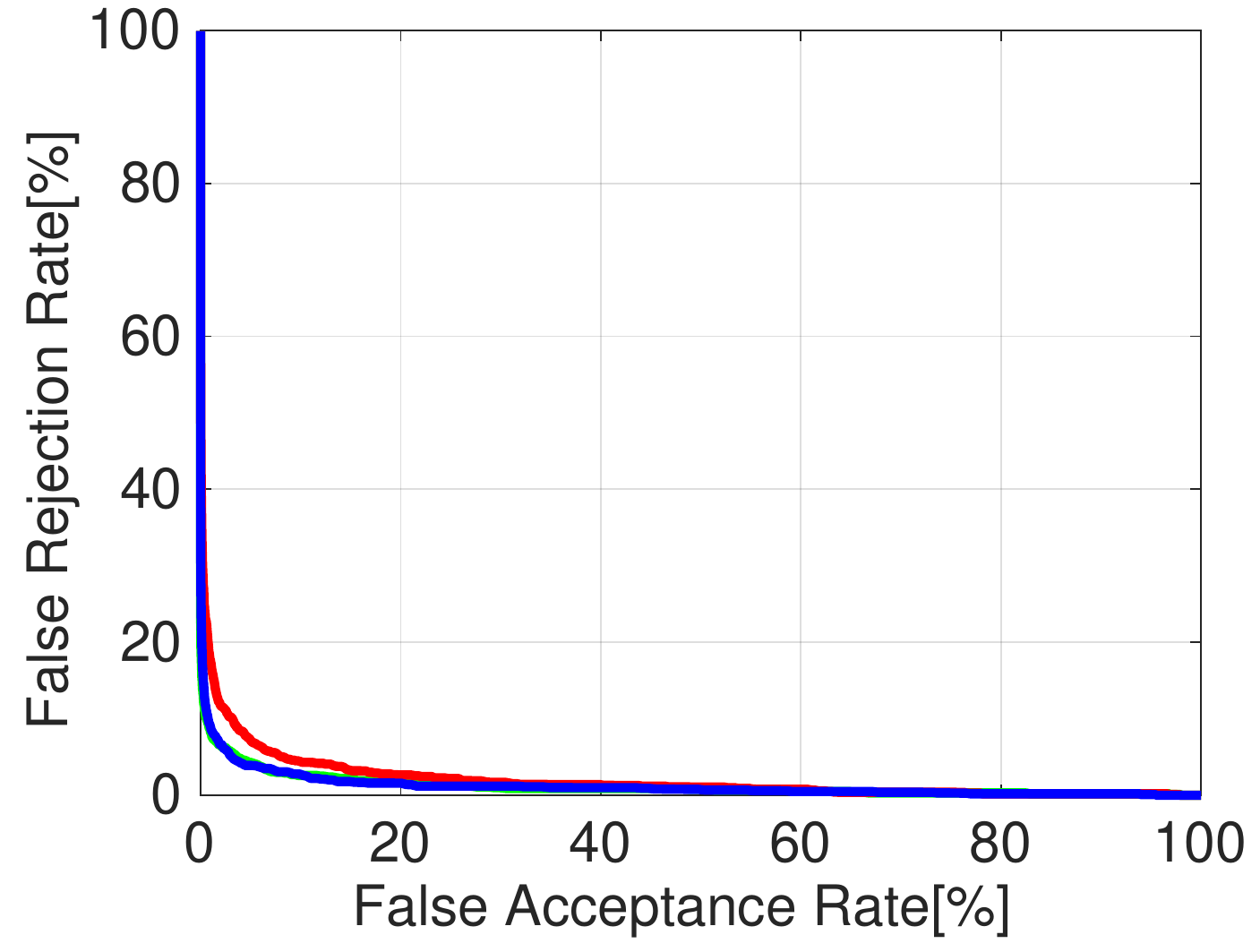} &
		\hspace{-3mm}
		\includegraphics[height=3.3cm]{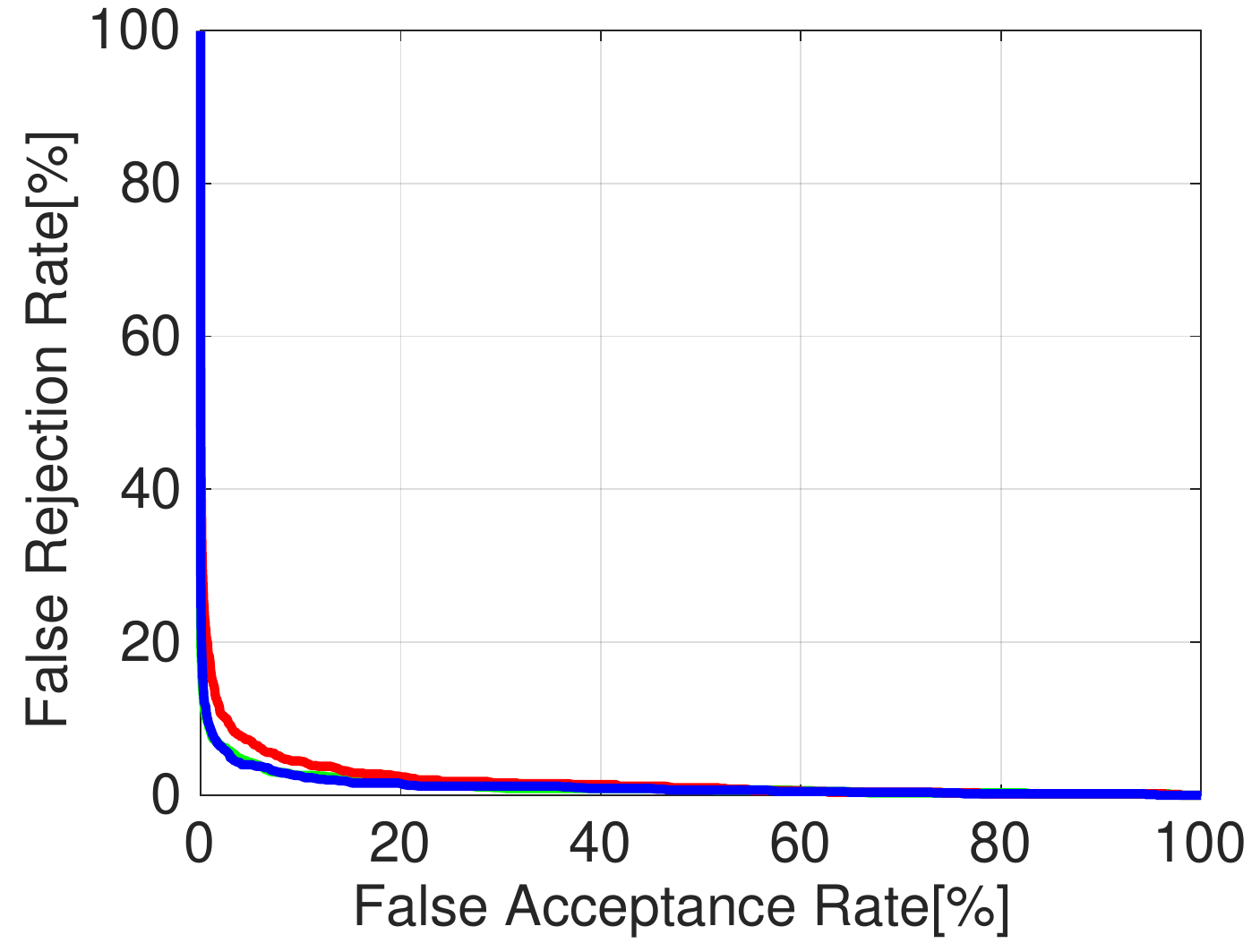} \\
    (g) 15 frames & (h) 18 frames & (i) 20 frames\\
		
\end{tabular}
\caption{Verification results on the OULP test data; The ROCs of the reconstructed GEIs (blue), incomplete GEIs (red), and true GEIs (green) for different number of frames.}
\label{fig:roc}
\end{figure}

In verification, first the similarity between two data samples (GEIs) is computed based on Euclidean distance. Then if the similarity score of this specific pair is less than or equal to the presupposed threshold, then this pair will be regarded as the positive pair, which means the GEIs in this pair belong to the same subject. Otherwise, it shows a negative pair, which indicates the GEIs in this pair do not belong to the same person.
The recognition results in verification tasks are described by ROCs in Fig.~\ref{fig:roc}. The ROC curves of three verification cases including IC-GEI, RC-GEI, and TC-GEI are presented for different number of frames. Although, as the number of frames $(N)$ composing the IC-GEIs gradually increases, the ROC of the IC-GEI get closer to the target ROCs, but there is still a gap between them. While, the ROCs of the reconstructed GEIs is relatively close to the target ROC even $1$ frame is used as the input. Note that as $N$ increases, the ROCs of the reconstructed GEIs gradually fit the target ROCs. 

The EERs calculated by these ROCs are listed in Table~\ref{tab:eer-oulp}. The left side of the table presents the EERs calculated for the reconstructed GEIs, and the right side shows the EERs computed for the incomplete GEIs. It is obvious that the reconstructed GEIs restored using IC-GEIs composed of more frames, have better performance. For instance, the EER of the IC-GEI composed of $20$ frames is $4.18\%$, which is very close to the target value $4.06\%$. In contrast to the left side of the table, the performance of IC-GEIs without restoration is inferior. For example, the EER of 1f-GEIs is $40.16\%$, which means almost half of the subjects are falsely matched. For IC-GEIs composed of more frames, their performance on verification tasks has been significantly improved after the restoration by the ITCNet. 

\begin{table}[!h]
\caption{Comparison of EERs [\%] of the verification test on the \textbf{OULP} dataset for different number of frames. The $n$f-RC-GEIs represent the RC-GEIs restored from $n$f-GEIs. The EER value for verification using true GEI (TC-GEI) is \textbf{4.06}.}
\centering
  \begin{tabular}{cc|cc}
  \hline
	\hline
Reconstructed GEI & EERs [\%]  & Incomplete GEI & EERs [\%]   \\
  \hline
1f-RC-GEIs   & 9.35  & 1f-GEIs & 40.16\\
3f-RC-GEIs   & 7.31  & 3f-GEIs & 37.97\\
5f-RC-GEIs   & 6.28  & 5f-GEIs & 34.61 \\
8f-RC-GEIs   & 5.52  & 8f-GEIs & 26.54  \\
10f-RC-GEIs  & 5.20  & 10f-GEIs & 18.49  \\
13f-RC-GEIs  & 4.75  & 13f-GEIs & 9.18  \\
15f-RC-GEIs  & 4.56  & 15f-GEIs & 7.39 \\
18f-RC-GEIs  & 4.43  & 18f-GEIs  & 6.20  \\
20f-RC-GEIs  & 4.18  & 20f-GEIs  & 6.12  \\
  \hline
  \end{tabular}
	\label{tab:eer-oulp}
\end{table}

\begin{table}[!h]
\caption{Comparison of EERs [\%] of the verification test on the \textbf{Casia-B} dataset for different number of frames. The $n$f-RC-GEIs represent the RC-GEIs restored from $n$f-GEIs. The EER value for verification using true GEI (TC-GEI) is \textbf{5.55}.}
\centering
  \begin{tabular}{cc|cc}
  \hline
	\hline
Reconstructed GEI & EERs [\%]  & Incomplete GEI & EERs [\%]   \\
  \hline
1f-RC-GEIs   & 22.77  &  1f-GEIs  &  39.44 \\
2f-RC-GEIs   & 20.83  &  2f-GEIs  &  35.57 \\
4f-RC-GEIs   & 19.44  &  4f-GEIs  &  39.23    \\
6f-RC-GEIs   & 19.16  &  6f-GEIs  &  35.57  \\
8f-RC-GEIs   & 15.01  &  8f-GEIs  &  35.07  \\
10f-RC-GEIs  & 14.16  &  10f-GEIs &  30.01  \\
13f-RC-GEIs  & 11.11  &  13f-GEIs &  23.60  \\
15f-RC-GEIs  & 10.55  &  15f-GEIs &  19.99  \\
17f-RC-GEIs  & 10.27  &  17f-GEIs &  19.44 \\
  \hline
  \end{tabular}
	\label{tab:eer-casia}
\end{table}

\section{Conclusion}
\label{sec:conclusion}
We have proposed a fully convolutional neural network for gait energy image (GEI) reconstruction from an incomplete gait cycle. The model could reconstruct a GEI, given an incomplete-GEI which is composed of only a few frames of a gait cycle. This model can successfully reconstruct a true GEI from an incomplete GEI, despite the starting frame and the number of available frames. Experimental results on two large gait datasets show that the proposed model can improve recognition rate greatly, particularity when there is only $0.1$ part of a gait cycle is available. 
In future, we will extend this model to an end-to-end model for both gait energy image reconstruction and recognition. Having one end-to-end deep model for both GEI reconstruction and recognition is also interesting.

\bibliography{mybibfile}

\end{document}